\documentclass{article}

\usepackage{rgdr_arxiv,times}

\usepackage{amsmath,amsfonts,bm}

\def\eqref#1{equation~\ref{#1}}

\def\1{\bm{1}}

\DeclareMathAlphabet{\mathsfit}{\encodingdefault}{\sfdefault}{m}{sl}
\SetMathAlphabet{\mathsfit}{bold}{\encodingdefault}{\sfdefault}{bx}{n}

\usepackage{amsmath}
\usepackage{amssymb}
\usepackage{amsthm}
\usepackage{booktabs}
\usepackage{graphicx}
\usepackage{placeins}
\usepackage{afterpage}
\usepackage{needspace}
\usepackage[table]{xcolor}
\usepackage{float}
\usepackage{algorithm}
\usepackage{algpseudocode}
\algrenewcommand\algorithmicrequire{\textbf{Input}}
\algrenewcommand\algorithmicensure{\textbf{Output}}
\newenvironment{ldsalgorithm}[2][t]{%
  \begin{algorithm}[#1]%
  \caption{#2}%
}{%
  \end{algorithm}%
}

\usepackage{hyperref}
\usepackage{url}
\definecolor{citationblue}{RGB}{0,0,128}
\hypersetup{
  colorlinks=true,
  citecolor=citationblue,
  linkcolor=black,
  urlcolor=citationblue,
  pdftitle={Reflection-Guided Dynamics Rectification for Text-to-Image Generation}
}

\theoremstyle{plain}
\newtheorem{theorem}{Theorem}[section]

\theoremstyle{definition}

\title{Reflection-Guided Dynamics Rectification for Text-to-Image Generation}

\author{}
\rgdrfinalcopy

\makeatletter
\newcommand{\LDSInlineMainTable}[1]{%
  \begingroup
  \renewenvironment{table*}[1][]{\def\@captype{table}}{}%
  \input{#1}%
  \endgroup}
\makeatother
\newsavebox{\LDSMainTableBox}
\newsavebox{\LDSMainCaptionBox}

\begin{document}
\raggedbottom
\makeatletter
\renewcommand{\@maketitle}{%
  \newpage\null
  \begin{center}%
    {\LARGE \bfseries \@title \par}%
  \end{center}%
  \vskip 0.6em 
}
\makeatother

\maketitle
\begin{center}
  \large
  Shuo~Xu\textsuperscript{1,2$*$},
  Hanrui~Ying\textsuperscript{1},
  Debin~Liu\textsuperscript{3},
  Jianjun~Li\textsuperscript{1},
  Zhiyuan~Ma\textsuperscript{1$\dagger$}
  
  \vspace{1.2em}
  
  \footnotesize
  \textsuperscript{1}Huazhong University of Science and Technology \quad
  \textsuperscript{2}Sorbonne University \quad
  \textsuperscript{3}Zhengzhou University

  \vspace{0.4em}
  \textsuperscript{$*$}Project leader and first author.\quad
  \textsuperscript{$\dagger$}Corresponding author.
\end{center}

\begingroup
  \renewcommand{\thefootnote}{}
  \makeatletter
  \let\@footnotetext\H@@footnotetext
  \renewcommand{\@makefntext}[1]{\noindent #1}
  \makeatother
  \footnotetext{\urlstyle{same}Interactive demos, dataset, and code for this paper are available at: \url{https://mair-lab-hust.github.io/RGDR/}.}
\endgroup

\fancyhead{}
\lhead{Preprint}
\rhead{Preprint}

\begin{center}
  \textit{Preprint.}
\end{center}
\vspace{0.5em}

\fancyhead{}
\lhead{Preprint}
\rhead{Preprint}

\vspace{0.5em}
\raggedbottom

\begin{figure}[!htbp]
  \centering
  \includegraphics[width=\textwidth]{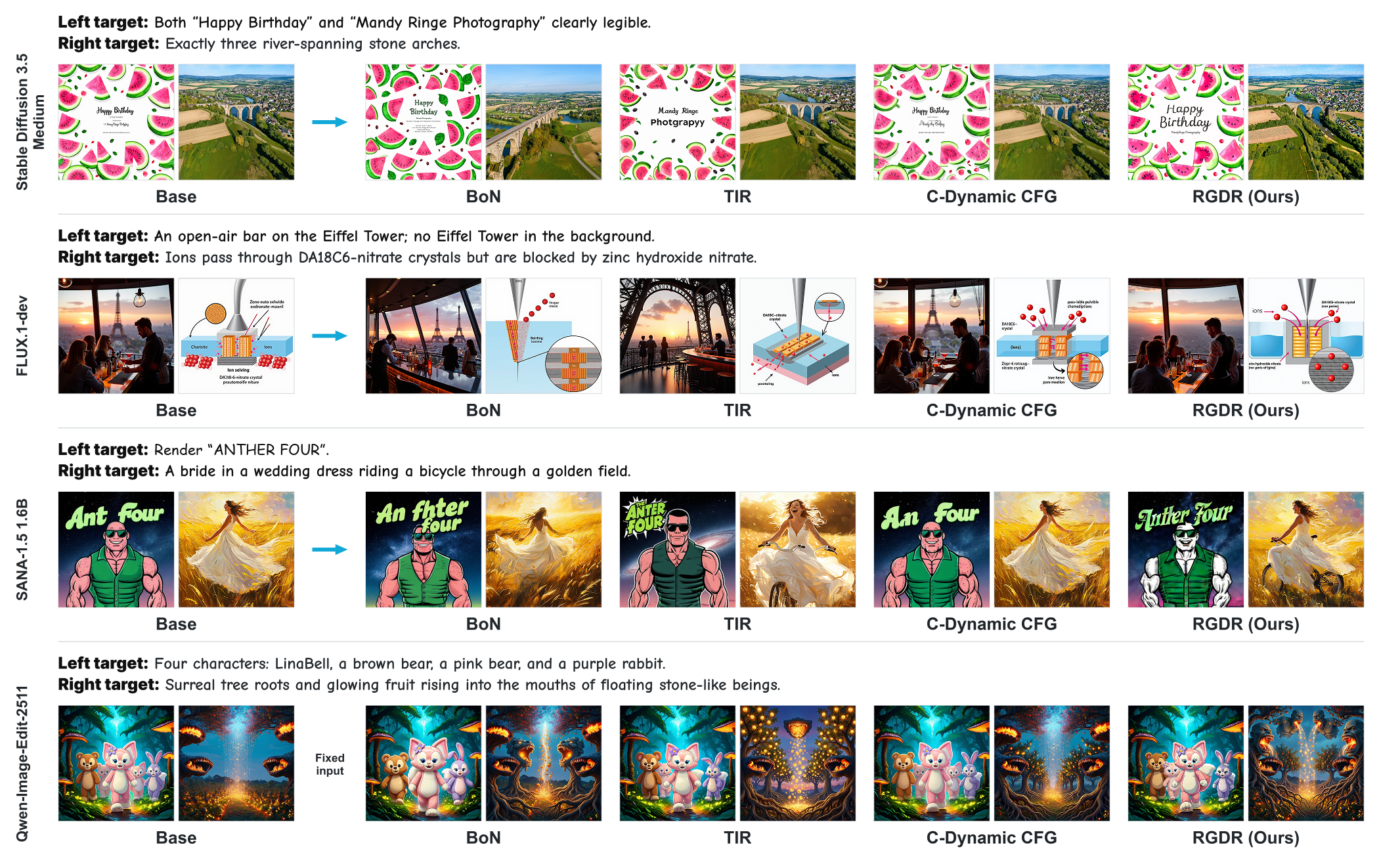}
  \caption{\textbf{Reflection-guided Dynamics Rectification (RGDR).} RGDR turns reflections on existing outputs into stepwise corrections along sampling trajectories. Qualitative comparisons with BoN, TIR, and C-Dynamic CFG across one image-editing model and three text-to-image models illustrate how different methods address residual errors and affect existing content during rectification.}
  \label{fig:main-qualitative}
\end{figure}

\begin{abstract}

Fine-grained rectification, which corrects residual errors in outputs from initial optimization, remains a key challenge in test-time optimization (TTO) for image generation, particularly on tasks requiring complex reasoning. In this setting, even when reflection repeatedly points out these errors, models may fail to correct them or even disrupt existing structure, revealing an execution gap between reflection and correction. This gap is particularly difficult to bridge with continued use of classical approaches such as prompt refinement or noise sampling, which treat generation as a black box, leaving the resulting trajectories and outputs uncontrolled. To narrow this gap, we propose Reflection-Guided Dynamics Rectification (RGDR), an effective and controllable paradigm for fine-grained rectification after initial TTO in flow models. This training-free method incorporates semantic guidance into TTO's feedback loop, turning reflection into direct corrective interventions in sampling dynamics. Moreover, RGDR supports continuous adjustment of intervention strength for fine-grained control over generation trajectories. Experiments across text-to-image and image-editing models show that the Target--Reference construction yields the largest rectification gain among the tested variants. Averaged equally across four models, RGDR achieves $1.51$--$2.68\times$ the rectification gain of the compared baselines and the highest gain per VLM token.\looseness=-1

\end{abstract}

\section{Introduction}

\label{sec:introduction}


Research on text-to-image generation is increasingly directed toward achieving higher image quality and better alignment with prompts. Specifically, as inference costs accumulate, test-time optimization (TTO) demonstrates its superiority without updating the parameters of a pretrained model. On tasks requiring complex reasoning, prior methods have made progress in correcting errors such as incorrect object types or counts through sampling, search, and feedback-driven refinement. They use generative models to produce additional candidates or image-editing models to revise the output~\citep{wan-etal-2026-test,jiang2026raise}. However, fine-grained rectification of the errors that remain after such optimization is still a key challenge under a limited additional budget.\looseness=-1


Two common forms of test-time scaling for image generation operate at the noise and prompt levels~\citep{zhuo2025reflectionflow}. At the noise level, best-of-$N$ (BoN) sampling generates candidates from different initial noises under a fixed prompt and selects the highest-scoring output~\citep{ma2025scaling}. At the prompt level, test-time prompt refinement (TIR) iteratively revises the prompt using feedback on generated images~\citep{khan2025tir}. Although these approaches remain effective, further adjustments to noise or prompts cannot sustain increasing gains under a limited budget. Figure~\ref{fig:intro-limitations} contrasts Noise Sampling and Prompt Refinement. The former produces substantial visual variation, whereas the latter repeatedly produces similar outputs, yet key requirements remain unmet in the illustrated examples. Across evaluated settings, the mean score gain per additional image also generally decreases as the budget grows. These observations indicate an execution gap between reflection and correction. We believe that this gap is mainly due to the black-box treatment of generation. These approaches adjust inputs and evaluate outputs without examining intermediate states. Error formation is therefore not explicitly tracked. It also remains unclear whether input adjustments steer the trajectory in the intended direction with an appropriate strength. These observations suggest that TTO methods for fine-grained rectification should place greater emphasis on image generation trajectories, where deviations can be identified for effective and controllable correction of residual errors.\looseness=-1

\afterpage{%
\begin{figure}[t]
  \centering
  \includegraphics[width=\textwidth,trim=0 10bp 0 0,clip]{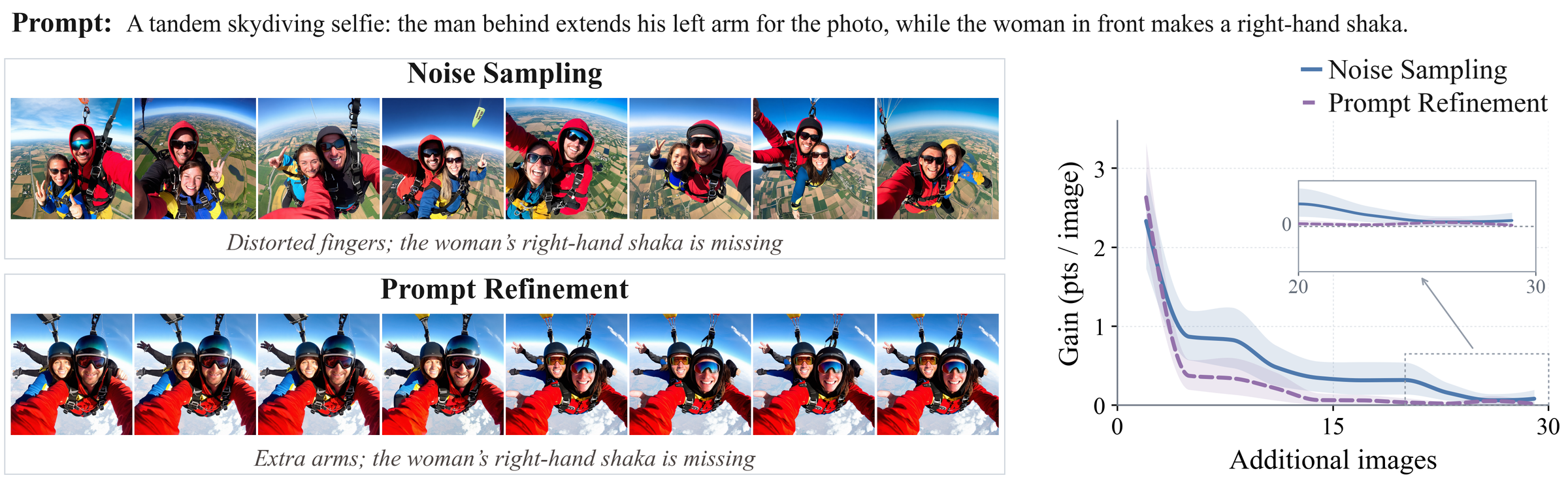}
  \caption{\textbf{Limitations of noise sampling and prompt refinement in fine-grained rectification.} The left panel shows qualitative results from noise sampling (top) and prompt refinement (bottom). The right panel reports mean score gain per additional image across generation budgets.}
  \label{fig:intro-limitations}
\end{figure}
\afterpage{%
\begin{figure}[t]
  \centering
  \includegraphics[width=\textwidth]{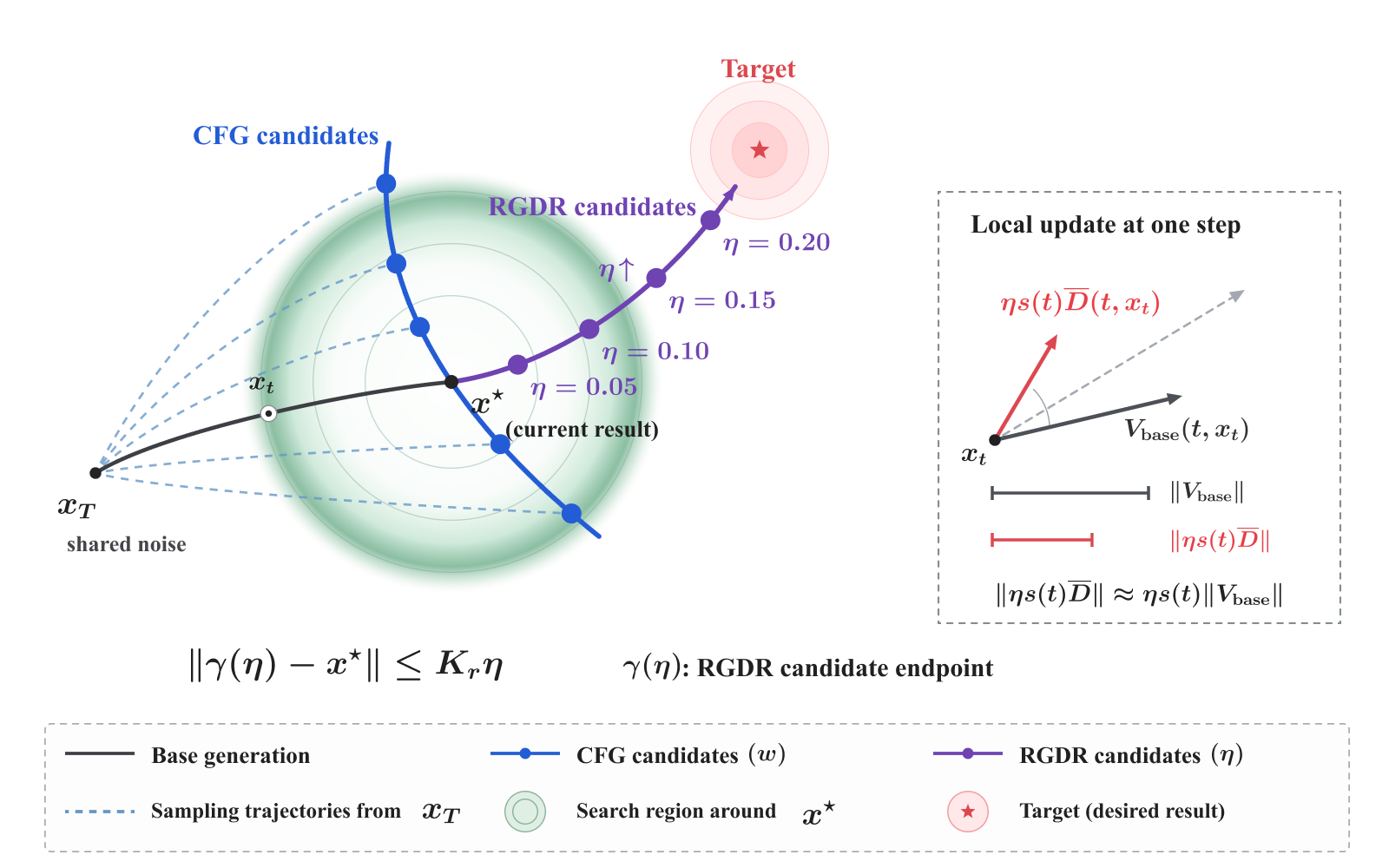}
  \caption{\textbf{Continuously controlled candidate space.} The blue and purple curves represent candidate endpoints parameterized by the CFG scale $w$ and RGDR relative strength $\eta$.}
  \label{fig:lds-cfg-candidate-spaces}
\end{figure}
\afterpage{%
\begin{figure}[t]
  \centering
  \includegraphics[width=\textwidth]{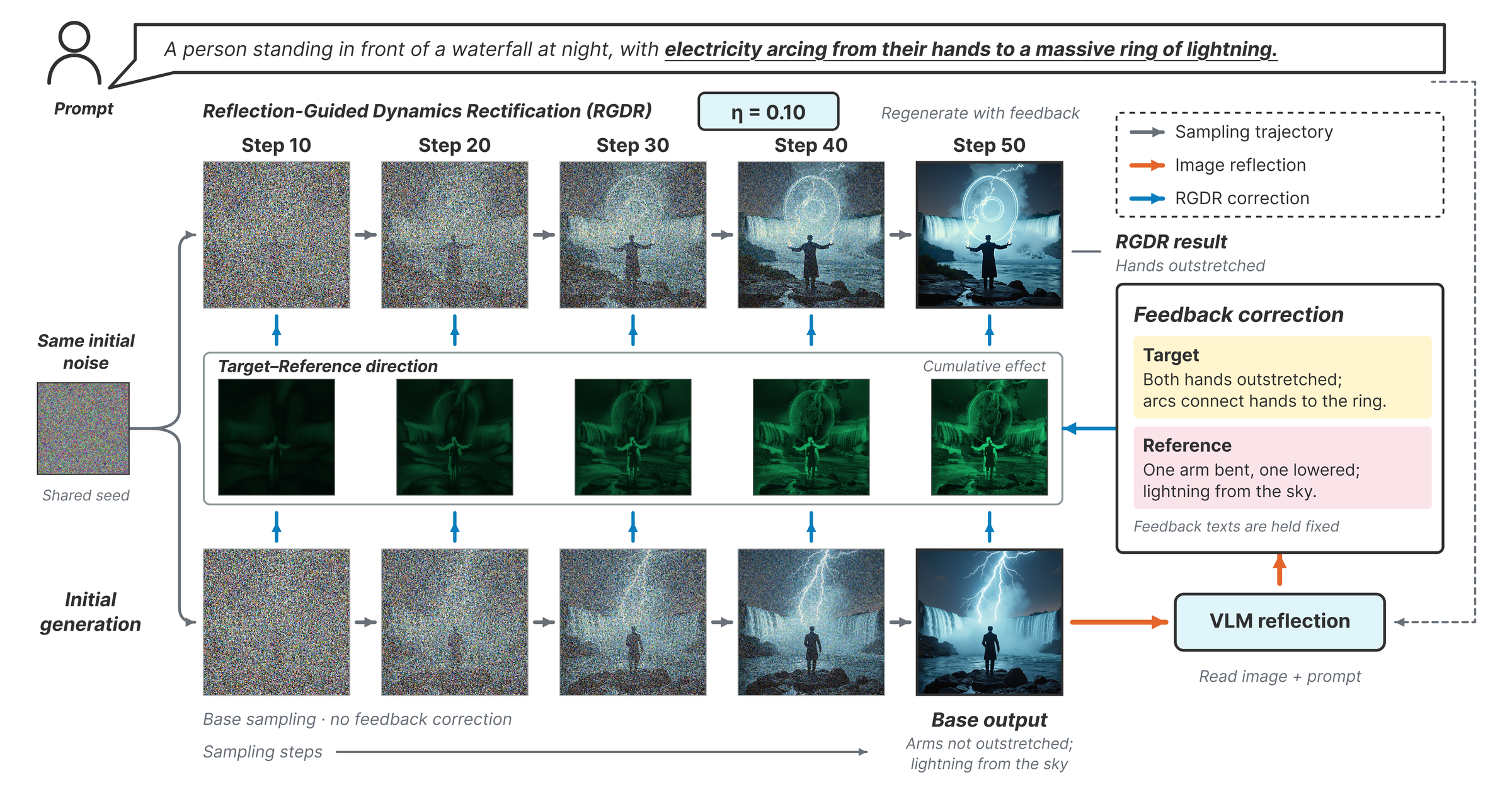}
  \caption{\textbf{Stepwise visualization with RGDR.} Reflections on the completed reference provide fixed Target--Reference conditions for constructing the correction field. Heatmaps show the spatial distribution of the correction signal along the generation trajectory.}
  \label{fig:lds-stepwise-correction}
\end{figure}
}%
}%
}%


In terms of previous work, evaluations or diagnoses of intermediate states guide the following steps along generation trajectories. For example, Dynamic CFG selects guidance scales via online evaluation, while PPAD revises semantic conditions and corrects through local rollback and denoising~\citep{papalampidi2025dynamiccfg,lv2025ppad}. These methods use feedback to select generation settings or update semantic conditions, exploring alternative trajectories for better outputs. Although the CFG framework provides a continuous control space for image generation~\citep{CFG}, feedback-driven trajectory exploration alone does not necessarily bridge the execution gap between reflection and correction. As illustrated in Figure~\ref{fig:lds-cfg-candidate-spaces}, varying the CFG scale under similar prompts and the same initial noise can still yield a constrained family of candidate endpoints. When the desired corrected output lies outside this family, which means that the model struggles to achieve the required rectification through its own capabilities alone, further search over these scales cannot resolve the error. Crucially, a key requirement for bridging this gap and extending the model's rectification capability is to directly steer sampling trajectories with a targeted correction signal derived from existing reflection. Such guidance redirects trajectory exploration toward regions closer to the desired corrected outputs, enabling more effective exploration.\looseness=-1


Given the above analysis, we introduce Reflection-Guided Dynamics Rectification (RGDR), an effective and controllable paradigm for fine-grained rectification after initial TTO in flow models. RGDR incorporates semantic composition into TTO's feedback loop, adding a reflection-guided correction field to the base velocity at each step. Varying correction strength just by adjusting the coefficients from the CFG framework produces gradually changing candidates under the same reflection, with continuity established under suitable regularity conditions. Figure~\ref{fig:lds-stepwise-correction} compares trajectories from the same initial noise to show how this intervention affects sampling at different time steps. Further theoretical analysis guides us to target the attributes, objects, or relations requiring correction and reduce interference with already-satisfied requirements. Under this criterion, we draw on positive and negative semantic composition~\citep{liu2022composable,brack2023sega} to construct a Target--Reference correction field from the difference between velocity predictions conditioned on the desired and current erroneous states. Our direction ablations show this construction achieves the largest rectification gain among tested alternatives.\looseness=-1


We compare RGDR with multiple baselines~\citep{ma2025scaling,khan2025tir,papalampidi2025dynamiccfg} for fine-grained rectification on complex reasoning tasks drawn from PRISM-Bench~\citep{fang2025prism} and SciIR-Bench~\citep{ma2026sciir}. Our evaluation spans commonly used text-to-image and image-editing models~\citep{esser2024sd3,bfl2024flux,xie2025sana15,wu2025qwenimagetechnicalreport}. We assess rectification gains against inference costs measured by wall-clock time and external-model token usage. Experiments show that, after initial optimization, RGDR achieves the highest rectification gains among the compared methods, with higher VLM-token efficiency and time efficiency comparable to C-Dynamic CFG. Further analysis shows that closing the correction window before sampling ends largely preserves image quality, offering a practical direction for reducing generation time and improving time efficiency.\looseness=-1

\section{Background and Related Work}
\label{sec:background}
\subsection{Flow-Based Generation}

We formulate trajectory rectification in flow models as an intervention on the sampling velocity field. Flow models describe continuous transformations between noise and data distributions through time-dependent velocity fields. Flow matching~\citep{FlowMatching} and rectified flow~\citep{liu2023flow} learn these fields and generate samples using numerical ordinary differential equation (ODE) solvers. Under the time convention from the noise endpoint $T$ to the data endpoint $0$, a sampling trajectory conditioned on $c$ satisfies
\begin{equation}
\frac{\mathrm d x_t}{\mathrm dt}
=
V(t,x_t;c),
\qquad
x_T\sim\mathcal N(0,I_d),
\qquad
t:T\rightarrow0.
\label{eq:generative-ode}
\end{equation}
Here, $x_t\in\mathbb R^d$ is the generative state, $T>0$ is the sampling horizon, $I_d$ is the $d$-dimensional identity matrix, and $V(t,x;c)\in\mathbb R^d$ is the velocity field incorporating the chosen guidance configuration. The states $x_T$ and $x_0$ denote the initial noise and generated endpoint, respectively. In latent-space models, a decoder maps $x_0$ to the final image~\citep{Rombach_2022_CVPR}. This framework supports both text-to-image generation and image editing. For example, $\text{FLUX.1 Kontext}$~\citep{bfl2025kontext} unifies image generation and conditional editing within a flow-matching model.

\subsection{Semantic Guidance and Control}
\label{sec:additive-background}

Semantic guidance combines or modifies model predictions, steering generation trajectories by changing sampling updates. Classifier-free guidance (CFG)~\citep{CFG} combines conditional and unconditional predictions and scales their difference. Composable Diffusion~\citep{liu2022composable} combines conditional predictions to implement concept conjunction and negation. SEGA~\citep{brack2023sega} supplements prompt guidance with independently adjustable sparse concept directions, allowing additional semantics to influence sampling updates. ANSWER~\citep{desai2025answer} estimates negative guidance during sampling without explicit negative prompts or an external captioning model. CFG-Ctrl~\citep{wang2026cfgctrl} interprets guidance as feedback control over sampling dynamics and develops a sliding-mode controller to steer generation trajectories.\looseness=-1

\subsection{Test-Time Optimization for Image Generation}

Test-time optimization improves image generation through additional sampling, search, and feedback-driven refinement~\citep{wan-etal-2026-test}. BoN~\citep{ma2025scaling} selects candidates from different noise initializations, while TIR~\citep{khan2025tir} refines prompts using generated images and feedback. ReflectionFlow~\citep{zhuo2025reflectionflow} and Reflect-DiT~\citep{li2025reflectdit} train models to revise outputs using images and textual reflections, whereas RAISE~\citep{jiang2026raise} coordinates prompt rewriting, noise resampling, and image editing without additional training. At the trajectory level, related work uses evaluations and diagnoses of intermediate states to guide subsequent sampling. Search over Paths~\citep{ma2025scaling} selects branches using denoised predictions, while Dynamic CFG~\citep{papalampidi2025dynamiccfg} uses latent-space evaluation to select guidance scales at each step. PPAD~\citep{lv2025ppad} diagnoses intermediate previews and revises denoising through corrective conditions and rollback. Extended discussion of these directions appears in Appendix~\ref{app:extended-related-work}.

\section{Method}
\label{sec:method-redraft}

RGDR introduces reflection-guided corrections into the sampling dynamics through an additive velocity field
\begin{equation}
\mathrm d x_t
=
\bigl[V_{\mathrm{base}}(t,x_t)+\alpha D(t,x_t)\bigr]\,\mathrm dt,
\qquad t:T\rightarrow0,
\label{eq:background-additive-field}
\end{equation}
where $V_{\mathrm{base}}(t,x_t)$ is the base velocity field, $D(t,x_t)\in\mathbb R^d$ is an additional semantic direction field, and $\alpha$ is a scaling coefficient. Both fields are evaluated at the same time and current state. Effective and controllable rectification requires addressing two challenges.
\begin{itemize}
  \setlength{\itemsep}{2pt}
  \setlength{\parsep}{0pt}
  \item \textbf{Correction direction.} Developing a consistent construction for translating diverse semantic feedback from TTO into correction direction fields that target the diagnosed errors.
  \item \textbf{Intervention strength.} Establishing a unified parameterization to control the influence of these fields on sampling trajectories across different reflections for the same model.
\end{itemize}
For controllable rectification, we first establish that candidate endpoints vary continuously with intervention strength and introduce a relative strength parameterization (Section~\ref{sec:controlled-candidates-redraft}). For effective rectification, we then analyze the conditions for corrective progress to guide the semantic construction of the correction field (Section~\ref{sec:feedback-correction-redraft}).\looseness=-1

\subsection{Control of Intervention Strength}
\label{sec:controlled-candidates-redraft}

\paragraph{Continuity and bounded deviation of the candidate family.} Fix the base field, correction field, and initial noise. Let $\gamma_D(\alpha)$ denote the endpoint obtained by solving Eq.~(\ref{eq:background-additive-field}) from $T$ to $0$. A strength range $\mathcal A$ defines the candidate family $\Gamma_D(\mathcal A)=\{\gamma_D(\alpha):\alpha\in\mathcal A\}$, with reference endpoint $x^\star=\gamma_D(0)$.\looseness=-1

\begin{theorem}[Lipschitz continuity of the ODE endpoint map]
\label{thm:method-controlled-candidates}
Fix $r>0$. Suppose that the dynamics admit a unique solution for every $\alpha\in[0,r]$. On a common region containing the relevant trajectories, assume that $V_{\mathrm{base}}$ and $D$ are continuous in time and uniformly Lipschitz in the state, and that $D$ is bounded. Then there exists a finite constant $K_r$, independent of $\alpha,\alpha'$, such that, for all $\alpha,\alpha'\in[0,r]$,
\begin{equation}
\|\gamma_D(\alpha)-\gamma_D(\alpha')\|\le K_r|\alpha-\alpha'|,\qquad \|\gamma_D(\alpha)-x^\star\|\le K_r\alpha.
\label{eq:method-endpoint-control}
\end{equation}
\end{theorem}
The two bounds describe continuous dependence of the endpoint on strength and its deviation from the reference, respectively. The proof is provided in Appendix~\ref{app:proof-continuity}.

\paragraph{Relative reparameterization of the intervention strength.} The theorem characterizes endpoint variation for a fixed direction field. The relative norms of the correction and base fields vary with feedback and sampling state, so the same raw coefficient can produce different relative intervention strengths. We therefore reparameterize the additional term relative to the base velocity norm,
\begin{equation}
\begin{aligned}\overline D(t,x)&=\frac{\|V_{\mathrm{base}}(t,x)\|}{\|D(t,x)\|+\varepsilon}D(t,x),\qquad\varepsilon>0,\\ V_{\eta,s}(t,x)&=V_{\mathrm{base}}(t,x)+\eta s(t)\overline D(t,x).\end{aligned}
\label{eq:method-lds-normalization}
\end{equation}
Here, $\eta\ge0$ sets the overall relative strength and $s(t)\ge0$ is a fixed, bounded temporal schedule. By construction,
\begin{equation}
\|V_{\eta,s}(t,x)-V_{\mathrm{base}}(t,x)\|\le\eta s(t)\|V_{\mathrm{base}}(t,x)\|.
\label{eq:method-relative-norm-bound}
\end{equation}
The correction at each sampling step is thus scaled relative to the current base velocity, with $\eta s(t)$ bounding its relative norm. This gives $\eta$ a dimensionless interpretation as a relative intervention parameter, rather than merely a multiplier of the raw direction. The schedule $s(t)$ distributes the correction strength along the generation trajectory and serves as an algorithm hyperparameter.

\begin{figure}[t]
  \centering
  \includegraphics[width=\textwidth,trim=0 16bp 0 12bp,clip]{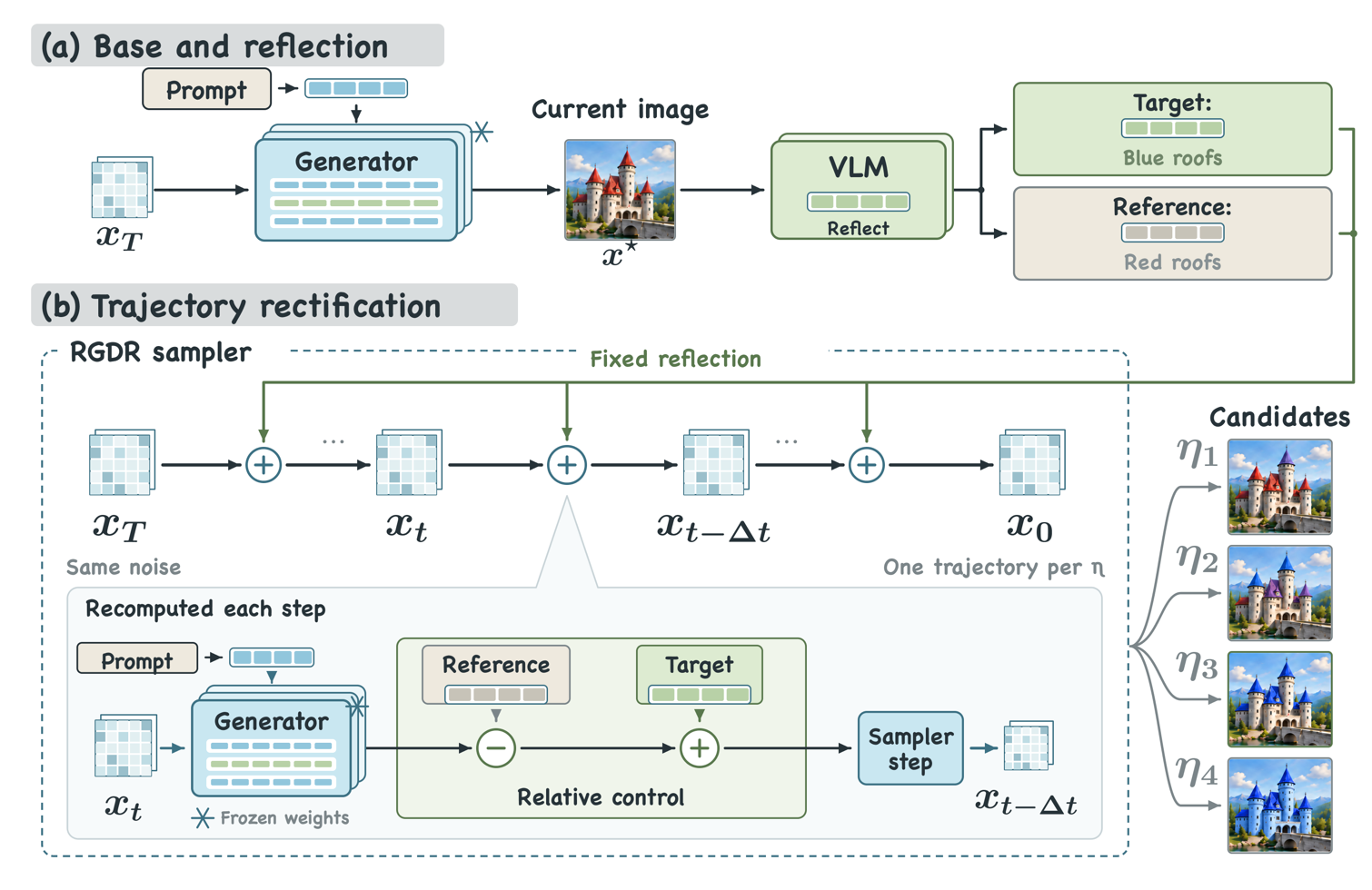}
  \setlength{\abovecaptionskip}{4pt}
  \makeatletter
  \long\def\@makecaption#1#2{\vskip\abovecaptionskip\noindent\hbox to\hsize{#1: #2\hfil}\par\vskip\belowcaptionskip}
  \makeatother
  \caption{\textbf{RGDR pipeline overview.}}
  \label{fig:lds-method-flow}
\end{figure}

RGDR therefore intervenes in the base sampling dynamics with the correction field $s(t)\overline D(t,x)$, scaled by the relative strength $\eta$. With the direction field, schedule, and initial noise fixed, Theorem~\ref{thm:method-controlled-candidates} establishes that the candidate endpoints $\gamma(\eta)$ vary continuously with $\eta$ under suitable regularity conditions, with bounded deviations from the reference. Figure~\ref{fig:lds-cfg-candidate-spaces} schematically illustrates this continuous candidate space. Figure~\ref{fig:lds-method-flow} shows a concrete example, where strengths $\eta_1$ to $\eta_4$ produce candidates with gradually changing roof colors.

\subsection{Semantic Construction of the Correction Field}
\label{sec:feedback-correction-redraft}

\textbf{Corrective progress and accompanying interference.} Fix the task $q$, feedback $F$, and the base field, initial noise, and schedule $s$ from Section~\ref{sec:controlled-candidates-redraft}. We write $D_F$ for the direction field $D$ constructed from $F$, and retain $\gamma(\eta)$ for the endpoint under relative reparameterization. Let $\Phi_q:\mathbb R^d\rightarrow\mathbb R^{m_q}$ be a fixed task representation used for analysis, where $m_q$ is its dimension. For latent-space generation, this mapping includes image decoding. Define
\begin{equation}
z(\eta)=\Phi_q\!\left(\gamma(\eta)\right),
\qquad
\Delta z(\eta)=z(\eta)-z(0).
\label{eq:method-task-response}
\end{equation}
In this representation space, let the unit vector $e_{\mathrm{fix}}\in\mathbb R^{m_q}$ specify the desired correction direction, and let $P_{\mathrm{keep}}\in\mathbb R^{m_q\times m_q}$ be the orthogonal projector onto the local subspace sensitive to properties already satisfied by the reference, with $P_{\mathrm{keep}}^\top=P_{\mathrm{keep}}^2=P_{\mathrm{keep}}$. Let $\beta>0$ weight the interference term. These analytical objects remain fixed for the current correction and are assumed to capture the progress and interference of interest. Define the net corrective progress
\begin{equation}
\Delta_q(\eta)
=
\langle e_{\mathrm{fix}},\Delta z(\eta)\rangle
-
\beta\|P_{\mathrm{keep}}\Delta z(\eta)\|.
\label{eq:method-repair-interference}
\end{equation}
The first term measures signed progress along the desired direction; the second measures the magnitude of changes in the subspace associated with already satisfied properties. A positive $\Delta_q(\eta)$ means that corrective progress outweighs weighted accompanying interference.

\begin{theorem}[Dominance of corrective progress induced by the feedback direction]
\label{thm:method-correction-dominance}
Let $r>0$. Suppose that $z$ is twice continuously differentiable on $[0,r]$, with
\begin{equation}
\sup_{\eta\in[0,r]}\|z''(\eta)\|\le M<\infty.
\label{eq:method-task-curvature}
\end{equation}
If the task response induced by $D_F$ satisfies
\begin{equation}
\kappa_q(D_F)
:=
\langle e_{\mathrm{fix}},z'(0^+)\rangle
-
\beta\|P_{\mathrm{keep}}z'(0^+)\|
>0,
\label{eq:method-feature-condition}
\end{equation}
where $z'(0^+)$ is the right derivative at zero strength, then, for every $\eta\in[0,r]$,
\begin{equation}
\Delta_q(\eta)
\ge
\kappa_q(D_F)\eta
-
\frac{(1+\beta)M}{2}\eta^2.
\label{eq:method-feature-net-gain}
\end{equation}
Consequently, when $M>0$, we have $\Delta_q(\eta)>0$ for all strengths satisfying
\begin{equation}
0<\eta<
\min\!\left\{r,\frac{2\kappa_q(D_F)}{(1+\beta)M}\right\}.
\label{eq:method-positive-correction-interval}
\end{equation}
When $M=0$, the same conclusion holds for all $0<\eta\le r$.

\end{theorem}

This result provides a design criterion for the correction field. Its induced response should advance the desired correction while reducing interference with properties already satisfied by the reference. When this response satisfies $\kappa_q(D_F)>0$, a sufficiently small positive intervention yields positive net corrective progress. Together with Theorem~\ref{thm:method-controlled-candidates}, this motivates constructing correction fields around the attributes, objects, or relations requiring correction and continuously adjusting their strength. The analysis guides correction-field design, while the effectiveness and relative merits of specific constructions are evaluated empirically. The proof and the dynamical expansion of the first-order response are provided in Appendices~\ref{app:proof-correction} and~\ref{app:first-order-response}.

\textbf{Target--Reference semantic construction.} Guided by the design criterion above, we extract from feedback $F$ two conditions describing the desired state, $c_{\mathrm{target}}$, and the current erroneous state, $c_{\mathrm{reference}}$, of the same objects, attributes, or relations. Using the frozen model's conditional velocity predictions $V_\theta$, we construct
\begin{equation}
D_F(t,x)
=
V_\theta(t,x;c_{\mathrm{target}})
-
V_\theta(t,x;c_{\mathrm{reference}}).
\label{eq:method-target-reference}
\end{equation}
Both conditions focus on the attributes, objects, or relations to be corrected and retain the necessary context without restating unrelated requirements that are already satisfied. This design combines error-focused descriptions with a Target--Reference difference, aiming to increase the first-order margin between corrective progress and interference. Our direction ablations in Section~\ref{sec:direction-ablation} examine this difference formulation and the specificity and scope of the conditioning texts, and show that this construction achieves the largest rectification gain among the tested constructions.

\subsection{RGDR Algorithm}
\label{sec:finite-search-redraft}

\begin{samepage}
\begin{ldsalgorithm}[t]{Reflection-Guided Dynamics Rectification (RGDR)}
\label{alg:lds-redraft}
\begin{algorithmic}[1]
\Require Task $q$; reference endpoint $x^\star$; fixed initial noise $x_T$;
\Statex base field $V_{\mathrm{base}}$; frozen conditional field $V_\theta$; feedback $F$; schedule $s$;
\Statex relative strength set $\mathcal E$; stabilizer $\varepsilon>0$;
\Statex evaluator $\widehat S_{\mathrm{sel}}(q,x)$ scoring the image corresponding to endpoint $x$
\Ensure Selected endpoint $\widehat x$
\State Extract Target and Reference conditions from $F$
\State $D_F(t,x)\gets V_\theta(t,x;c_{\mathrm{target}})-V_\theta(t,x;c_{\mathrm{reference}})$
\Statex Evaluate both terms at the same $(t,x)$
\State $\overline D_F\gets\|V_{\mathrm{base}}\|D_F/(\|D_F\|+\varepsilon)$
\State $\mathcal P\gets\{(0,x^\star)\}$
\ForAll{$\eta\in\mathcal E$}
\State Restart from the fixed initial noise $x_T$
\State Integrate $V_{\mathrm{base}}+\eta s(t)\overline D_F$ over the full interval from $T$ to $0$
\Statex Recompute the fields at each current state
\State $\mathcal P\gets\mathcal P\cup\{(\eta,x_0)\}$
\EndFor
\State $(\widehat\eta,\widehat x)\in\operatorname*{arg\,max}_{(\eta,x)\in\mathcal P}\widehat S_{\mathrm{sel}}(q,x)$
\State \Return $\widehat x$
\end{algorithmic}
\end{ldsalgorithm}
\end{samepage}

\begin{table*}[t]
\centering
\setlength{\abovecaptionskip}{0pt}
\setlength{\belowcaptionskip}{6pt}
\caption{\textbf{Comparison of fine-grained rectification on our evaluation set from PRISM-Bench and SciIR-Bench.} Within each model, methods share starting images and candidate counts. Time and Token denote Gain per unit of generation time and token consumption.}
\label{tab:main-pooled}
\begingroup
\small
\setlength{\tabcolsep}{2pt}
\setlength{\arrayrulewidth}{0.3pt}
\renewcommand{\arraystretch}{1.00}
\setlength{\aboverulesep}{1pt}
\setlength{\belowrulesep}{1.5pt}
\begin{tabular*}{\linewidth}{@{\extracolsep{\fill}}lrrrr|rr@{}}
\toprule
 & \multicolumn{4}{c|}{Task Performance} & \multicolumn{2}{c}{Cost Efficiency} \\
\cmidrule(lr){2-5}\cmidrule(l){6-7}
Method & \shortstack[r]{Gain $\uparrow$\\(points)} & \shortstack[r]{Improved $\uparrow$\\(\%)} & \shortstack[r]{Repair $\uparrow$\\(\%)} & \shortstack[r]{Preserve $\uparrow$\\(\%)} & \shortstack[r]{Time $\uparrow$\\(points/s)} & \shortstack[r]{Token $\uparrow$\\(points/$10^4$ tokens)} \\
\midrule
\rowcolor{black!7}
\multicolumn{5}{>{\columncolor{black!7}[0pt][10pt]}l}{\textit{Stable Diffusion $3.5$ Medium}} & \multicolumn{2}{r}{\textit{Base = $56.90$}} \\
BoN & $+1.35$ & $12.0$ & $4.11$ & $96.14$ & $0.0351$ & $0.455$ \\
TIR & $+1.24$ & $13.5$ & $6.04$ & $\mathbf{98.39}$ & $0.0318$ & $0.213$ \\
C-Dynamic CFG & $+2.15$ & $23.0$ & $6.99$ & $97.87$ & $\mathbf{0.0553}$ & $0.686$ \\
\textbf{RGDR (Ours)} & $\mathbf{+3.74}$ & $\mathbf{28.0}$ & $\mathbf{12.96}$ & $96.44$ & $0.0469$ & $\mathbf{1.060}$ \\
\midrule
\rowcolor{black!7}
\multicolumn{5}{>{\columncolor{black!7}[0pt][10pt]}l}{\textit{SANA-$1.5$~$1.6$B}} & \multicolumn{2}{r}{\textit{Base = $46.55$}} \\
BoN & $+1.54$ & $13.0$ & $3.40$ & $\mathbf{99.08}$ & $0.1768$ & $0.514$ \\
TIR & $+3.41$ & $25.0$ & $9.71$ & $96.61$ & $\mathbf{0.3607}$ & $0.560$ \\
C-Dynamic CFG & $+2.42$ & $22.5$ & $6.55$ & $99.07$ & $0.2610$ & $0.738$ \\
\textbf{RGDR (Ours)} & $\mathbf{+4.37}$ & $\mathbf{31.5}$ & $\mathbf{13.07}$ & $94.46$ & $0.2591$ & $\mathbf{1.259}$ \\
\midrule
\rowcolor{black!7}
\multicolumn{5}{>{\columncolor{black!7}[0pt][10pt]}l}{\textit{FLUX.$1$-dev}} & \multicolumn{2}{r}{\textit{Base = $70.30$}} \\
BoN & $+1.27$ & $11.5$ & $7.92$ & $97.65$ & $0.0112$ & $0.449$ \\
TIR & $+2.99$ & $15.5$ & $10.23$ & $\mathbf{97.86}$ & $\mathbf{0.0264}$ & $0.563$ \\
C-Dynamic CFG & $+1.63$ & $16.5$ & $8.78$ & $97.65$ & $0.0144$ & $0.579$ \\
\textbf{RGDR (Ours)} & $\mathbf{+4.36}$ & $\mathbf{26.5}$ & $\mathbf{19.15}$ & $96.78$ & $0.0130$ & $\mathbf{1.379}$ \\
\midrule
\rowcolor{black!7}
\multicolumn{5}{>{\columncolor{black!7}[0pt][10pt]}l}{\textit{Qwen-Image-Edit-$2511$}} & \multicolumn{2}{r}{\textit{Base = $72.77$}} \\
BoN & $+2.07$ & $24.55$ & $13.99$ & $97.53$ & $0.0072$ & $1.264$ \\
TIR & $+3.41$ & $30.91$ & $20.32$ & $97.25$ & $\mathbf{0.0118}$ & $0.771$ \\
C-Dynamic CFG & $+3.03$ & $29.09$ & $16.27$ & $\mathbf{97.95}$ & $0.0105$ & $1.846$ \\
\textbf{RGDR (Ours)} & $\mathbf{+4.23}$ & $\mathbf{33.64}$ & $\mathbf{23.58}$ & $96.48$ & $0.0074$ & $\mathbf{2.078}$ \\
\bottomrule
\end{tabular*}
\endgroup
\end{table*}

Combining the controlled superposition in Section~\ref{sec:controlled-candidates-redraft} with the semantic construction in Section~\ref{sec:feedback-correction-redraft}, RGDR derives Target and Reference conditions from reflections on a completed reference output. These describe the desired and observed states of the elements requiring correction, respectively, retaining relevant context without requesting changes to already-correct content. These conditions remain fixed across sampling steps and candidate strengths. At each sampling step, RGDR recomputes the difference between the corresponding conditional velocity predictions, applies relative reparameterization, and adds the resulting correction to the base velocity field. With the generation configuration, feedback, and schedule fixed, candidates at different relative strengths are sampled separately from the same initial noise and evaluated alongside the reference for final selection. If no candidate receives a higher evaluation score, the reference can be retained. Model parameters remain fixed throughout. Algorithm~\ref{alg:lds-redraft} details the procedure, and Figure~\ref{fig:lds-method-flow} illustrates the pipeline.\looseness=-1


\section{Experiments}
\label{sec:experiments}

\subsection{Setup}
\label{sec:experimental-setup}

To evaluate fine-grained rectification of residual errors after TTO in both text-to-image generation and image editing, we build an evaluation set from PRISM-Bench and SciIR-Bench, covering compositional requirements and scientific image reasoning \citep{fang2025prism,ma2026sciir}. Each task pairs a prompt with starting images that satisfy most requirements but retain specific errors after initial optimization with conventional TTO methods, with seeds and sampling settings provided for reproduction. These images are generated by $\text{SD3.5 Medium}$, $\text{SANA-1.5 1.6B}$, and $\text{FLUX.1-dev}$, or obtained by editing $\text{SD3.5 Medium}$ outputs with $\text{Qwen-Image-Edit-2511}$. All parameters remain frozen. $\text{Seed-2.1 Pro}$ and $\text{GPT-5.6 Sol}$ supply feedback and score outputs against fixed checklists on a $0$--$100$ scale. We report Gain and Improved for rectification performance, Repair and Preserve for error correction and retention, and score gain per unit of generation time and token consumption for cost efficiency. Details of the main experiments and ablation studies are provided in Appendix~\ref{app:experiment-details}. Temporal-scheduling definitions and settings are provided in Appendix~\ref{app:temporal-scheduling}.

\subsection{Main Experiments}
\label{sec:experimental-results}

\paragraph{Comparison of TTO methods for fine-grained rectification.} Starting from the fixed images in our rectification evaluation set, we compare continued BoN sampling~\citep{ma2025scaling}, TIR prompt refinement~\citep{khan2025tir}, Constant Dynamic CFG (C-Dynamic CFG), our constant-schedule adaptation of Dynamic CFG~\citep{papalampidi2025dynamiccfg}, and RGDR across three text-to-image models and one image-editing model. BoN provides a baseline for gains from random sampling and best-of-N selection, allowing us to assess whether targeted rectification yields further improvements beyond this process. Within each model, all methods share the same starting images and candidate counts, with the starting image also retained for final selection. Comparison protocols and cost accounting are detailed in Appendix~\ref{app:comparison-protocols}.

\begin{figure}[t]
  \centering
\begin{table*}[t]
\centering
\setlength{\abovecaptionskip}{0pt}
\setlength{\belowcaptionskip}{6pt}
\caption{\textbf{Effects of semantic construction and reflection quality on RGDR.} Reflection style describes the specificity and scope of the feedback. Gain is measured in score points.}
\label{tab:direction-ablation}
\begingroup
\small
\setlength{\tabcolsep}{1.5pt}
\setlength{\arrayrulewidth}{0.3pt}
\renewcommand{\arraystretch}{1.00}
\begin{tabular*}{\textwidth}{@{\extracolsep{\fill}}llrrrr@{}}
\toprule
\multicolumn{6}{@{}l@{}}{\textbf{Control baselines:}\quad Gaussian: $1.31$\qquad TR Wrong: $1.15$} \\
\midrule
Feedback VLM & Reflection style & TR Correct & Target Only & Local Operation & Global Operation \\
\midrule
$\text{Seed-2.1 Pro}$ & Error-focused & $\mathbf{4.34}$ & $2.32$ & $3.34$ & $2.60$ \\
$\text{GPT-5.6 Sol}$ & Global and abstract & $\mathbf{2.86}$ & $2.41$ & $1.83$ & $2.25$ \\
\bottomrule
\end{tabular*}
\endgroup
\end{table*}

  \par\vspace{2pt}
  \includegraphics[width=\textwidth]{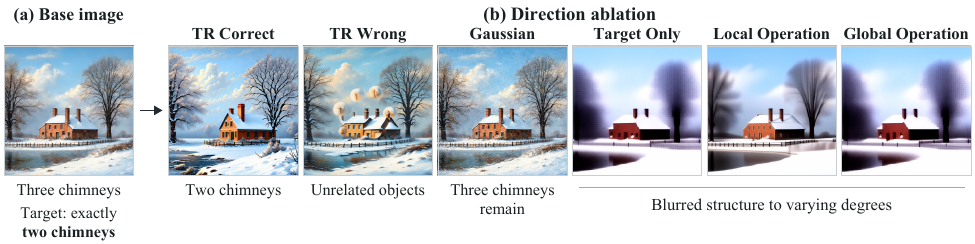}
  \caption{\textbf{Qualitative comparison of semantic constructions on SD3.5 Medium.} (a) Starting image with three chimneys; the target is exactly two. (b) Selected outputs from four semantic constructions using focused $\text{Seed-2.1 Pro}$ feedback, alongside TR Wrong and Gaussian controls.}
  \label{fig:direction-ablation}
\end{figure}

Table~\ref{tab:main-pooled} shows that RGDR achieves the highest Gain, Improved, and Repair across all four models. Averaged equally across models, its Gain exceeds that of BoN and TIR by $2.62$ and $1.41$ points, respectively, and reaches $1.81\times$ that of C-Dynamic CFG. Under matched additional-image budgets, these results demonstrate more effective rectification than further noise sampling, prompt refinement, or native guidance-scale adjustment, supporting the error-specific additional direction illustrated in Figure~\ref{fig:lds-cfg-candidate-spaces}. Figure~\ref{fig:main-qualitative} further illustrates controllability through the bridal cycling example: both TIR and RGDR introduce the bicycle and riding action, but RGDR retains the side-view figure, wedding dress, and golden field, whereas TIR changes the subject's orientation and composition. RGDR thus realizes the requested correction while maintaining closer correspondence to the starting image. Regarding cost, RGDR achieves the highest gain per VLM token across all four models, while its gain per unit of generation time remains comparable to C-Dynamic CFG's, with greater overall rectification gains. By searching over $\eta$ under a fixed reflection, RGDR both controls the correction and reduces dependence on repeated external VLM rewriting, at the cost of additional sampling computation.\looseness=-1

\subsection{Ablation Studies}
\label{sec:direction-ablation}

\paragraph{Semantic descriptions and direction construction.} Guided by the design criterion for the correction field discussed in Section~\ref{sec:feedback-correction-redraft}, we examine semantic descriptions and direction construction in RGDR. The former concerns whether reflections specifically and accurately identify residual errors and desired changes, and whether descriptions include unrelated, already-satisfied content; the latter compares ways of converting these descriptions into correction directions. Table~\ref{tab:direction-ablation} compares four semantic constructions under different feedback conditions. Gaussian and TR Wrong serve as control baselines, reporting gains obtained through candidate selection under random perturbations and mismatched semantic directions, respectively. Details of the ablation subset, feedback examples, and constructions are provided in Appendix~\ref{app:vlm-prompts}.

\begin{figure}[t]
  \centering
  \includegraphics[width=\textwidth]{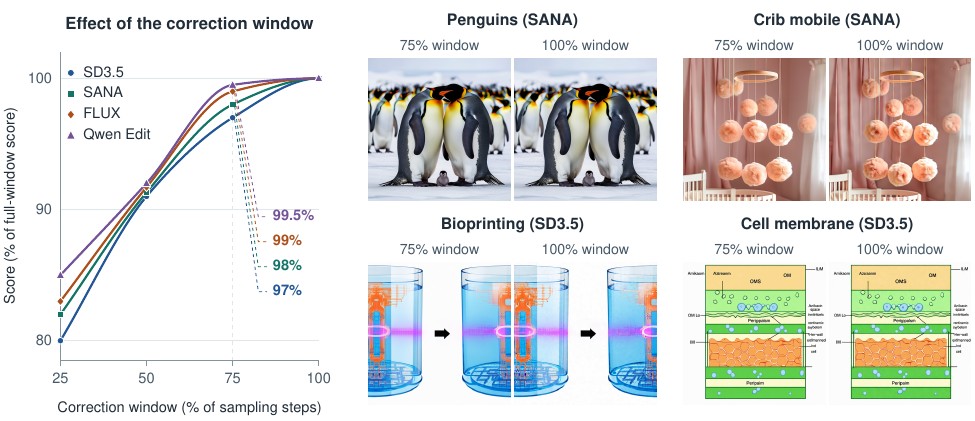}
  \caption{\textbf{Effect of the correction window.} Left: final mean score relative to full-window correction. Right: paired outputs using $75\%$ and $100\%$ windows on SANA (top) and $\text{SD3.5 Medium}$ (bottom). All sampling steps are completed.}
  \label{fig:correction-window}
\end{figure}

Table~\ref{tab:direction-ablation} provides empirical support for the design criterion of focusing on residual errors while limiting interference with already-correct content. Error-focused reflections generally yield larger gains than global, abstract feedback, with TR Correct improving from $2.86$ to $4.34$. TR Correct ranks first under both feedback conditions and outperforms Target Only, supporting the Target--Reference difference. All four constructions exceed the Gaussian and TR Wrong controls. Figure~\ref{fig:direction-ablation} illustrates these differences: TR Correct changes three chimneys to two; Gaussian mainly alters fine details while retaining the incorrect count; and TR Wrong introduces unrelated objects. Target Only and both Operation variants produce varying degrees of structural blur, but Local Operation causes less degradation than Global Operation in this example, consistent with the results under focused feedback in Table~\ref{tab:direction-ablation}. This suggests that focusing descriptions on the requested changes can mitigate interference even when the construction itself remains imperfect. Together, these results support the design criterion and Target--Reference as the most effective construction among those tested, while motivating exploration of further semantic constructions under the same criterion.\looseness=-1

\subsection{Further Discussion}
\label{sec:further-discussion}

RGDR requires additional conditional velocity predictions to construct the semantic correction field, increasing generation time. To explore ways to reduce this cost, we vary the correction window defined in Appendix~\ref{app:temporal-scheduling} in the RGDR ablation while completing all sampling steps. Applying correction only during the first $75\%$ of steps retains $97$--$99.5\%$ of the full-window mean score. Figure~\ref{fig:correction-window} shows the score trends across correction windows on the left and paired outputs with similar visual quality under the $75\%$ and full windows on the right. Skipping additional correction predictions outside the window provides a practical direction for reducing computation and improving time efficiency.\looseness=-1

\FloatBarrier

\section{Conclusion}
\label{sec:conclusion}

We address the rectification of residual errors after initial optimization and introduce RGDR as an effective and controllable paradigm for fine-grained rectification in flow models. RGDR integrates semantic guidance into TTO's feedback loop, holding reflection fixed while continuously adjusting correction strength for targeted trajectory interventions. This paradigm has two technical foundations: relative reparameterization for strength control, with continuity established under suitable conditions; and a theoretically motivated direction-construction criterion that focuses on errors while reducing interference with already-correct content. Our ablations support this criterion and identify Target--Reference as the most effective construction among those tested. Across four models, RGDR achieves the largest rectification gains and the highest gain per VLM token, with average time efficiency comparable to C-Dynamic CFG. Correction-window experiments further indicate scope to reduce additional computation while largely retaining image quality. Future work will explore more diverse semantic constructions and compatible mechanisms such as SEGA-style sparsification, together with more efficient intervention schedules for stable, lower-cost rectification.\looseness=-1

\section*{AI Use Statement}
We used generative AI tools to assist with manuscript polishing, the writing of mathematical proofs, the refinement of existing hypotheses, code development, and figure layout and composition. We did not use these tools to originate the core research ideas or claims. For schematic figures, AI assistance was limited to layout and composition; the authors determined the content and drew the figures themselves. AI-assisted proofs and hypothesis refinements were checked for logical consistency, code was reviewed and verified for correctness, and figures were inspected for accuracy and scientific plausibility. We take responsibility for the final content of this work.

\section*{REPRODUCIBILITY STATEMENT}
To make our results reproducible, we provide relevant details in the appendix, including theoretical proofs, task and checklist construction, model and sampling configurations, evaluation protocols, and implementation details of the proposed method.

\bibliography{references}
\bibliographystyle{rgdr_arxiv}

\clearpage
\appendix

\renewcommand*{\theHequation}{appendix.\thesection.\arabic{equation}}
\renewcommand*{\theHtable}{appendix.\thesection.\arabic{table}}

\section{Extended Related Work}
\label{app:extended-related-work}

\noindent\textbf{Input- and Output-Level Optimization.} Beyond the noise sampling and prompt refinement methods discussed in the main text, ReNO~\citep{eyring2024reno} optimizes initial noise using reward gradients. Idea2Img~\citep{yang2023idea2img} iteratively updates prompts based on diagnoses of generated images, while PRIS~\citep{kim2025pris} aggregates recurring errors across candidates to guide prompt redesign. Other work enhances multi-round reflection through training. SRRL~\citep{pan2025srrl} and RL-RIG~\citep{wang2026rlrig} use reinforcement learning to improve multi-round image generation and reflection-based editing, respectively.

\noindent\textbf{Trajectory-Level Optimization.} Feynman--Kac Steering~\citep{singhal2025fk} explores generation trajectories through particle resampling, whereas Classical Search~\citep{zhang2025inferencetimescalingdiffusionmodels} combines local and tree search to expand sampling paths. T2S2~\citep{sakurai2026t2s2} detects prohibited concepts in intermediate clean-image estimates, optimizes a low-rank conditioning residual, and restarts sampling.

\section{Mathematical Proofs}
\label{app:math-proofs}
\setcounter{equation}{0}
\renewcommand{\theequation}{\thesection.\arabic{equation}}

We retain the definitions and assumptions of the main text. Vector norms are Euclidean, and matrix norms are the corresponding induced operator norms.

\subsection{Proof of Theorem~\ref{thm:method-controlled-candidates}}
\label{app:proof-continuity}

Let $L_V,L_D$ be the uniform state-Lipschitz constants of the base and direction fields, respectively, and choose a finite $C_D\ge0$ such that $\|D(t,x)\|\le C_D$. For $\alpha,\alpha'\in[0,r]$, let $x_t^\alpha,x_t^{\alpha'}$ start from the same initial noise $x_T$. Define $e(t)=\|x_t^\alpha-x_t^{\alpha'}\|$, so that $e(T)=0$. Under the time convention $T\rightarrow0$, the integral equation is
\[
x_t^\alpha=x_T-\int_t^T\bigl[V_{\mathrm{base}}(u,x_u^\alpha)+\alpha D(u,x_u^\alpha)\bigr]\,\mathrm du.
\]
Subtracting the equations at $\alpha$ and $\alpha'$ and applying the triangle inequality gives
\[
\begin{aligned}
e(t)&\le\int_t^T\|V_{\mathrm{base}}(u,x_u^\alpha)-V_{\mathrm{base}}(u,x_u^{\alpha'})\|\,\mathrm du\\
&\quad+\int_t^T\|\alpha D(u,x_u^\alpha)-\alpha'D(u,x_u^{\alpha'})\|\,\mathrm du.
\end{aligned}
\]
The base-field difference is bounded by $L_Ve(u)$. To separate the effects of the state and coefficient in the correction term, write
\[
\begin{aligned}
&\alpha D(u,x_u^\alpha)-\alpha'D(u,x_u^{\alpha'})\\
&\quad=\alpha\bigl[D(u,x_u^\alpha)-D(u,x_u^{\alpha'})\bigr]
+(\alpha-\alpha')D(u,x_u^{\alpha'}).
\end{aligned}
\]
The Lipschitz and boundedness assumptions, together with $0\le\alpha\le r$, therefore give
\[
\begin{aligned}
\|\alpha D(u,x_u^\alpha)-\alpha'D(u,x_u^{\alpha'})\|
&\le\alpha L_De(u)+C_D|\alpha-\alpha'|\\
&\le rL_De(u)+C_D|\alpha-\alpha'|.
\end{aligned}
\]
Combining these estimates yields
\begin{equation}
e(t)\le\int_t^T\bigl[(L_V+rL_D)e(u)+C_D|\alpha-\alpha'|\bigr]\,\mathrm du.
\label{eq:app-trajectory-difference}
\end{equation}
Set $L_r=L_V+rL_D$ and introduce forward time $\tau=T-t$, with $h(\tau)=e(T-\tau)$. Changing variables in the integral gives
\[
h(\tau)\le\int_0^\tau\bigl[L_rh(v)+C_D|\alpha-\alpha'|\bigr]\,\mathrm dv.
\]
Applying Gr\"onwall's inequality \citep[Lemma~2.7]{teschl2012ordinary} with zero initial term and constant forcing $C_D|\alpha-\alpha'|$ gives
\[
h(\tau)\le C_D|\alpha-\alpha'|\int_0^\tau e^{L_r(\tau-v)}\,\mathrm dv.
\]
Returning to the original time variable, we obtain
\begin{equation}
e(t)\le C_D|\alpha-\alpha'|\int_t^T e^{L_r(u-t)}\,\mathrm du.
\label{eq:app-gronwall}
\end{equation}
Taking $t=0$ and defining $K_r=C_D\int_0^T e^{L_ru}\,\mathrm du<\infty$, we obtain
\begin{equation}
\|\gamma_D(\alpha)-\gamma_D(\alpha')\|\le K_r|\alpha-\alpha'|.
\label{eq:app-endpoint-lipschitz}
\end{equation}
Setting $\alpha'=0$ and using $x^\star=\gamma_D(0)$ then gives $\|\gamma_D(\alpha)-x^\star\|\le K_r\alpha$. This completes the proof. The integral expression for $K_r$ also covers $L_r=0$ and $C_D=0$.

\paragraph{Application after relative reparameterization.}Suppose additionally that $\|V_{\mathrm{base}}(t,x)\|\le C_V<\infty$ on the common trajectory region. The reverse triangle inequality and the $2/\varepsilon$-Lipschitz bound for $u\mapsto u/(\|u\|+\varepsilon)$ imply that the reparameterized direction in the main text satisfies
\begin{equation}
\begin{aligned}
\|\overline D(t,x)\|&\le C_V,\\
\|\overline D(t,x)-\overline D(t,y)\|&\le\left(L_V+\frac{2C_VL_D}{\varepsilon}\right)\|x-y\|.
\end{aligned}
\label{eq:app-normalized-regularity}
\end{equation}
Thus, for a fixed continuous bounded schedule $s$, the field $s\overline D$ satisfies the same types of boundedness and state-Lipschitz conditions, and the preceding comparison applies to $\gamma(\eta)$. The same integral estimate also holds for bounded piecewise-continuous schedules, provided the corresponding solutions exist.

\subsection{Proof of Theorem~\ref{thm:method-correction-dominance}}
\label{app:proof-correction}

Let $w=z'(0^+)$. Applying Taylor's formula with an integral remainder componentwise to $z$ \citep[Theorem~1.36]{knapp2016basic} gives
\begin{equation}
\Delta z(\eta)=\eta w+R(\eta),\qquad
R(\eta)=\int_0^\eta(\eta-u)z''(u)\,\mathrm du.
\label{eq:app-taylor}
\end{equation}
The bound on the second derivative implies
\[
\|R(\eta)\|\le\int_0^\eta(\eta-u)\|z''(u)\|\,\mathrm du
\le M\int_0^\eta(\eta-u)\,\mathrm du=\frac M2\eta^2.
\]
We estimate corrective progress and accompanying interference separately. Since $\|e_{\mathrm{fix}}\|=1$, the Cauchy--Schwarz inequality gives
\[
\begin{aligned}
\langle e_{\mathrm{fix}},\Delta z(\eta)\rangle
&=\eta\langle e_{\mathrm{fix}},w\rangle+\langle e_{\mathrm{fix}},R(\eta)\rangle\\
&\ge\eta\langle e_{\mathrm{fix}},w\rangle-\|R(\eta)\|.
\end{aligned}
\]
Because $P_{\mathrm{keep}}$ is an orthogonal projector, $\|P_{\mathrm{keep}}\|\le1$. The triangle inequality and $\eta\ge0$ yield
\[
\begin{aligned}
\|P_{\mathrm{keep}}\Delta z(\eta)\|
&\le\eta\|P_{\mathrm{keep}}w\|+\|P_{\mathrm{keep}}R(\eta)\|\\
&\le\eta\|P_{\mathrm{keep}}w\|+\|R(\eta)\|.
\end{aligned}
\]
Multiplication by $-\beta<0$ reverses this inequality, so
\[
-\beta\|P_{\mathrm{keep}}\Delta z(\eta)\|
\ge-\beta\eta\|P_{\mathrm{keep}}w\|-\beta\|R(\eta)\|.
\]
Combining the two estimates and substituting the remainder bound gives
\begin{equation}
\begin{aligned}
\Delta_q(\eta)
&=\langle e_{\mathrm{fix}},\Delta z(\eta)\rangle
-\beta\|P_{\mathrm{keep}}\Delta z(\eta)\|\\
&\ge\eta\bigl(\langle e_{\mathrm{fix}},w\rangle-\beta\|P_{\mathrm{keep}}w\|\bigr)
-(1+\beta)\|R(\eta)\|\\
&\ge\kappa_q(D_F)\eta-\frac{(1+\beta)M}{2}\eta^2.
\end{aligned}
\label{eq:app-correction-bound}
\end{equation}
When $\kappa_q(D_F)>0$ and $M>0$, this lower bound factors as
\[
\eta\left(\kappa_q(D_F)-\frac{(1+\beta)M}{2}\eta\right).
\]
Both factors are strictly positive for $0<\eta<2\kappa_q(D_F)/((1+\beta)M)$. Restricting to the interval of the theorem therefore gives $\Delta_q(\eta)>0$ whenever
\begin{equation}
0<\eta<\min\left\{r,\frac{2\kappa_q(D_F)}{(1+\beta)M}\right\}.
\label{eq:app-effective-interval}
\end{equation}
When $M=0$, we have $R(\eta)=0$, so $\Delta_q(\eta)=\kappa_q(D_F)\eta>0$ for every $0<\eta\le r$.

\subsection{Dynamical Expression of the First-Order Response}
\label{app:first-order-response}

To expand $z'(0^+)$ in the main text, additionally assume that $V_{\mathrm{base}}$ and $s\overline D_F$ are continuously differentiable in the state, with state derivatives bounded in a neighborhood of the relevant trajectories, and that $\Phi_q$ is differentiable at $x^\star$. The fields and their state derivatives are continuous in time. Under these conditions, differentiable dependence on parameters \citep[Theorem~9.16]{teschl2012ordinary} justifies differentiating the integral equation with respect to $\eta$. Let $x_t^0$ denote the reference trajectory and define
\begin{equation}
p_t=\left.\partial_\eta x_t^\eta\right|_{\eta=0^+},\qquad
A(t)=\mathrm D_xV_{\mathrm{base}}(t,x_t^0).
\label{eq:app-response-definition}
\end{equation}
Differentiating the sampling dynamics with respect to $\eta$ and evaluating at $\eta=0$ yields
\begin{equation}
\frac{\mathrm d p_t}{\mathrm dt}=A(t)p_t+s(t)\overline D_F(t,x_t^0),\qquad p_T=0.
\label{eq:app-variational}
\end{equation}
The condition $p_T=0$ follows from using the same initial noise at every strength. Let the state-transition matrix $\Psi(t,u)$ satisfy $\partial_t\Psi(t,u)=A(t)\Psi(t,u)$ and $\Psi(u,u)=I_d$. The variation-of-constants formula \citep[Theorem~3.12]{teschl2012ordinary}, followed by the chain rule, yields
\begin{equation}
\begin{aligned}
p_0&=\int_T^0\Psi(0,t)s(t)\overline D_F(t,x_t^0)\,\mathrm dt,\\
z'(0^+)&=\mathrm D\Phi_q(x^\star)\int_T^0\Psi(0,t)s(t)\overline D_F(t,x_t^0)\,\mathrm dt.
\end{aligned}
\label{eq:app-propagated-response}
\end{equation}
This expression propagates each step's correction through the subsequent dynamics to the generated endpoint and then maps it into the task representation. The integration direction agrees with the $T\rightarrow0$ convention in the main text. The additional smoothness assumptions in this subsection are used for this dynamical expansion, not for the preceding proof stated directly in terms of $z$.

\Needspace{16\baselineskip}
\section{Experimental Details}
\label{app:experiment-details}
\setcounter{equation}{0}
\renewcommand{\theequation}{\thesection.\arabic{equation}}
\setcounter{table}{0}
\renewcommand{\thetable}{S\arabic{table}}

\subsection{Evaluation Set for Fine-Grained Rectification}
\label{app:data-references}

We construct an evaluation set for fine-grained rectification. Each task includes a prompt, a checklist, and one starting image for each of the selected text-to-image and image-editing models. These images typically retain clear logical errors after conventional test-time optimization and serve as starting points for fine-grained rectification.

We organize a task library from PRISM-Bench and SciIR-Bench. To describe its content coverage, we organize tasks into three groups.

\begin{enumerate}
\item \textbf{Human.} This group covers human-centered images and requirements concerning pose, hand gestures, and interactions.
\item \textbf{Open.} This group covers general scenes and requirements concerning objects, attributes, spatial relations, and environmental conditions.
\item \textbf{Scientific \& Text.} This group combines scientific diagrams with text-containing images, including scientific structures and relations, diagram labels, and specified rendered text.
\end{enumerate}

The task library contains 1,500 entries, comprising 700 from PRISM-Bench and 800 from SciIR-Bench. Following the three groups defined above, it includes 176 Human tasks, 384 Open tasks, and 940 Scientific \& Text tasks. Each entry includes the original prompt, source information, and its associated checklist. For PRISM-Bench, we construct and review task-specific checklists from the original prompts to assess the requested visual content. For SciIR-Bench, we retain the officially provided checklist questions and categories. 

From this library, we curate a $400$-task evaluation set comprising $100$ Human, $100$ Open, and $200$ Scientific \& Text tasks. For each task and selected model configuration, we first use different random seeds to produce multiple candidates, either by generating images from the prompt or by editing an input image. We then select a starting image, prioritizing candidates that satisfy most requirements but still contain at least one clear logical error. Each record includes the original prompt, checklist, and model, along with the random seed and settings needed to reproduce the selected starting image. Section~\ref{app:model-implementation} lists the models and sampling configurations used for image generation and editing.

\subsection{Model Configuration}
\label{app:model-implementation}

Table~\ref{tab:app-model-settings} lists the model sampling configurations. Inference uses $\text{BF16}$ precision and a resolution of $1024\times1024$.

\begin{table}[htbp]
\centering\small
\caption{Model sampling configurations.}
\label{tab:app-model-settings}
\begin{tabular}{lrr}
\toprule
Model & Steps $N$ & Guidance scale\\
\midrule
$\text{SD3.5 Medium}$ &$40$&$4.5$\\
$\text{SANA-1.5 1.6B}$ &$20$&$4.5$\\
$\text{FLUX.1-dev}$ &$50$&$3.5$\\
$\text{Qwen-Image-Edit-2511}$ &$40$&$4.0$\\
\bottomrule
\end{tabular}
\end{table}

\subsection{Evaluation, Selection, and Aggregation}
\label{app:evaluation}

We evaluate images against fixed task-specific checklists and select final outputs under the same evaluation protocol. Let task $i$ have $K_i$ checklist items, and let $a_{ik}(I)\in[0,1]$ be the numerical score of image $I$ on item $k$. The task score is
\begin{equation}
S_i(I)=\frac{100}{K_i}\sum_{k=1}^{K_i}a_{ik}(I).
\label{eq:app-checklist-score}
\end{equation}
Let $b_i,f_i$ denote the scores of the shared reference and selected output, respectively, and let $n$ be the number of tasks in the aggregate. We report
\begin{equation}
\begin{aligned}
\mathrm{Final}&=\frac1n\sum_i f_i, &
\mathrm{Gain}&=\frac1n\sum_i(f_i-b_i),\\
\mathrm{Rel.\ gain}&=100\frac{\mathrm{Gain}}{\frac1n\sum_i b_i}, &
\mathrm{Improved}&=\frac{100}{n}\sum_i\mathbf 1\{f_i>b_i+\delta\}.
\end{aligned}
\label{eq:app-score-metrics}
\end{equation}
The recorded protocols for the three generative models use an improvement tolerance of $10^{-10}$ on the $[0,1]$ scale, equivalent to $\delta=10^{-8}$ on the $0$--$100$ scale. Rel. gain is the ratio of mean gain to mean reference score, not the mean of per-task relative percentages.

Repair and Preserve use paired pass/fail judgments. For task $i$, let $N_{ab}^{(i)}$ count items with reference judgment $a$ and output judgment $b$, where $0$ denotes fail and $1$ denotes pass. Items marked uncertain at either endpoint are excluded. Define $\mathcal I_R=\{i:N_{00}^{(i)}+N_{01}^{(i)}>0\}$ and $\mathcal I_P=\{i:N_{10}^{(i)}+N_{11}^{(i)}>0\}$. Then
\begin{equation}
\begin{aligned}
\mathrm{Repair}&=\frac{100}{|\mathcal I_R|}\sum_{i\in\mathcal I_R}\frac{N_{01}^{(i)}}{N_{00}^{(i)}+N_{01}^{(i)}},\\
\mathrm{Preserve}&=\frac{100}{|\mathcal I_P|}\sum_{i\in\mathcal I_P}\frac{N_{11}^{(i)}}{N_{10}^{(i)}+N_{11}^{(i)}}.
\end{aligned}
\label{eq:app-binary-metrics}
\end{equation}
Each metric is macro-averaged over tasks with a nonzero denominator. Numerical scores and binary judgments are recorded separately. An uncertain item can contribute its numerical score to $S_i$, but not to Repair or Preserve. Results are pooled across sources using the relevant valid-task sets; cross-model comparisons assign equal weight to each model. $\text{Seed-2.1 Pro}$ scored the images underlying the results in Tables~\ref{tab:main-pooled} and~\ref{tab:direction-ablation}.

\subsection{Main-Experiment Comparison Protocol}
\label{app:comparison-protocols}

All methods start from the fixed images in our rectification evaluation set. Each method receives an additional budget of ten complete images for each text-to-image model and five for the image-editing model. Tasks and checklists remain fixed across methods. For each method, we score the additional candidates using the same protocol and select the highest-scoring image from the starting image and all additional candidates as the final output.

\paragraph{BoN.} Following BoN candidate generation and selection~\citep{ma2025scaling}, we retain the original prompt or editing instruction, generate candidates from different initial noise, and select by checklist score.

\paragraph{TIR.} The prompt or editing instruction is revised iteratively using the current image and its evaluation feedback~\citep{khan2025tir}. The reference seed remains fixed. Each round produces a complete candidate under the revised condition, which informs the next reflection.

\begin{figure}[t]
  \centering
  \includegraphics[width=\textwidth]{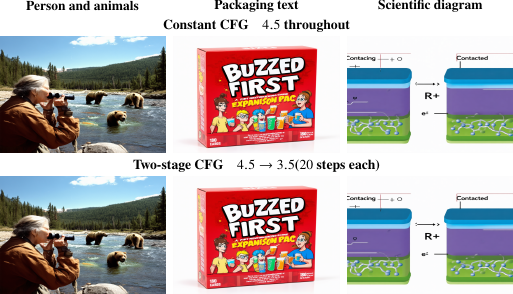}
  \caption{\textbf{Constant versus two-stage CFG on SD3.5 Medium.} Each column uses the same prompt and seed with $40$ sampling steps. Top: a constant scale of $4.5$. Bottom: $4.5$ for the first $20$ steps and $3.5$ for the remaining $20$ steps.}
  \label{fig:app-cfg-constant-two-stage}
\end{figure}

\paragraph{C-Dynamic CFG.} We use Constant Dynamic CFG (C-Dynamic CFG), a constant-schedule adaptation of Dynamic CFG~\citep{papalampidi2025dynamiccfg}. It retains the principle of using evaluation feedback to explore CFG guidance scales. The original method selects scales online using noisy-latent evaluators; our variant fixes $w_j\equiv w$ throughout each trajectory, generates complete candidates at different scales with the prompt and reference noise fixed, and evaluates them with the same VLM as other rectification methods. The main reason for this adaptation is that the original stepwise online search does not directly match our fine-grained rectification setting with a limited additional generation budget: varying guidance across sampling stages creates many possible schedules, while schedules differing only in later stages may yield similar results. In Figure~\ref{fig:app-cfg-constant-two-stage}, fixing the early-stage scale at $4.5$ and either retaining it or switching to $3.5$ in the second half produces highly similar outputs with persistent residual errors. We therefore vary the constant scale across candidates to explore different guidance settings from the start, prioritizing distinct early trajectories over later variations of similar trajectories and using the limited budget more effectively. $\text{SD3.5 Medium}$ and SANA use ten uniformly spaced guidance values in $[2.5,6.5]$; FLUX uses ten uniformly spaced native guidance-parameter values in $[2.1,4.9]$. For Qwen-Image-Edit-2511, we use $w\in\{1.5,2.5,3.5,5.0,6.0\}$, generating one complete candidate per value from the same fixed original input image.

\paragraph{RGDR.} Target--Reference conditions and reference noise remain fixed, and complete candidates are generated independently at different relative strengths. The text-to-image models use $\mathcal E=\{0.02,0.04,\ldots,0.20\}$ and the schedules in Appendix~\ref{app:temporal-scheduling}. The additional correction field is recomputed at each sampling step. For Qwen-Image-Edit-2511, we use $\eta\in\{0.01,0.025,0.05,0.075,0.10\}$ with constant scheduling, generating one complete candidate per value from the same fixed original input image.

\paragraph{Cost accounting.} We measure costs over the additional rectification stage, including all candidates and excluding the shared initial optimization. Generation time excludes VLM latency, while token usage includes input and output tokens for reflection or prompt rewriting and candidate evaluation. Time and Token report total score gain divided by total generation time and total token usage, respectively; Token is expressed in points per $10^4$ tokens. Table~\ref{tab:main-pooled} reports Qwen-Image-Edit-2511 results over the $330$ tasks that entered further rectification, excluding the $70$ of $400$ tasks whose starting images already satisfied all checklist requirements.

\FloatBarrier

\subsection{Ablation Directions and Conditioning Texts}
\label{app:vlm-prompts}
\suppressfloats[t]

\paragraph{Ablation subset.} We select $200$ tasks from the $400$-task rectification evaluation set for direction ablation, prioritizing cases where RGDR corrected some errors in the main experiments but clear errors remained. When all constructions produce no gain or reach full scores, their differences are difficult to observe. Cases with partial improvement and remaining errors allow us to compare whether different constructions correct more, about the same, or fewer errors from the same starting images.\looseness=-1

\paragraph{Conditioning texts.} We construct conditioning texts from the task requirements and residual errors in the starting image. Target describes the desired state after correction, while Reference describes the current erroneous state of the same objects, attributes, or relations. For error-focused Target and Reference descriptions, we retain the context needed to locate the correction without restating unrelated requirements already satisfied. We also examine broader descriptions that include additional scene content.

\paragraph{Feedback example.} The bridge example in Table~\ref{tab:app-feedback-example} illustrates differences in the specificity and scope of the conditioning texts produced by the two feedback VLMs.

\begin{table}[t]
\centering\small
\setlength{\tabcolsep}{3pt}
\renewcommand{\arraystretch}{1.08}
\caption{Target and Reference descriptions for the bridge example.}
\label{tab:app-feedback-example}
\begin{tabular}{@{}p{0.12\linewidth}p{0.41\linewidth}p{0.41\linewidth}@{}}
\toprule
Condition & $\text{Seed-2.1 Pro}$ & $\text{GPT-5.6 Sol}$ \\
\midrule
Reference & {\raggedright ``approximately ten large semicircular arches''\par} & {\raggedright ``The viaduct has many arches.''\par} \\
\midrule
Target & {\raggedright ``A prominent three-arched stone viaduct spanning the winding river, with three large stone arches forming the main span over the water.''\par} & {\raggedright ``a serene rural landscape with a prominent three-arched stone viaduct spanning a winding river a quaint village, expansive agricultural fields, and distant hills, viewed from the northeast.''\par} \\
\bottomrule
\end{tabular}
\end{table}

In this example, $\text{Seed-2.1 Pro}$ describes the current arch count more specifically and concentrates the Target description on the requested three-arch structure. $\text{GPT-5.6 Sol}$ describes the current count more broadly and includes additional scene content in its Target description.

\paragraph{Target--Reference (TR Correct).} We subtract the conditional velocity prediction under Reference from that under Target to construct the correction direction.

\paragraph{Target Only.} For each pair of Target--Reference texts, we use the same Target condition but omit the Reference-conditioned prediction when constructing the correction direction.

\paragraph{Local Operation.} We express the requested changes as a direct instruction focused on the errors to correct, retaining necessary context without adding descriptions of unrelated content to preserve. The velocity prediction conditioned on this instruction serves as the correction direction.

\paragraph{Global Operation.} We express the same requested changes as a direct instruction that also describes already-correct content to preserve. As in Local Operation, the instruction-conditioned velocity prediction serves as the correction direction. The distinction between the two Operations concerns the inclusion of preservation descriptions, while the specificity of the requested correction still depends on the reflection.

\paragraph{Control baselines.} Gaussian uses a random Gaussian direction without task semantics. TR Wrong retains the Target--Reference difference formulation but uses mismatched feedback. Their gains after candidate selection provide controls for random perturbations and mismatched semantic directions.

\paragraph{Comparison setting.} For a fair comparison, all constructions use relative reparameterization and the same number of candidates, with strength limits determined by preliminary visual checks to avoid severe image degradation. For example, $\text{SD3.5 Medium}$ uses an upper limit of $\eta=0.20$ for Target--Reference and $\eta=0.075$ for Global Operation.

\section{Temporal Scheduling}
\label{app:temporal-scheduling}
\label{app:supplementary-results}
\setcounter{equation}{0}
\renewcommand{\theequation}{\thesection.\arabic{equation}}

The temporal schedule separates the shape of correction strength from the window in which correction is active. For sampling steps indexed in execution order by $j=0,\ldots,N-1$, we write
\begin{equation}
s(t_j)=\bar{s}(t_j)\mathbf{1}_{\mathcal W}(t_j),
\label{eq:app-window-schedule}
\end{equation}
where $\bar{s}$ specifies the schedule shape and $\mathbf{1}_{\mathcal W}$ indicates whether the current step lies within the correction window. Common choices are
\begin{equation}
\bar{s}_{\mathrm{const}}(t_j)=1,\qquad
\bar{s}_{\mathrm{inc}}(t_j)=\frac{j}{N-1},\qquad
\bar{s}_{\mathrm{dec}}(t_j)=1-\frac{j}{N-1}.
\label{eq:app-discrete-schedule}
\end{equation}
Based on qualitative inspection, we use the increasing shape for $\text{SD3.5 Medium}$ and the constant shape for the other three models. The default setting enables correction throughout sampling. A window covering only the first $m$ steps sets $\mathcal W=\{t_0,\ldots,t_{m-1}\}$. Outside this window, the additional correction predictions are skipped while base sampling continues to completion. This formulation allows the schedule shape and correction window to be varied separately, with $\eta$ retaining its role as the overall relative correction strength.

\makeatletter
\newcommand{\appqualrow}[1]{%
  \par\noindent\includegraphics[width=\linewidth]{#1}\par\vspace{3pt}%
}
\newcommand{\appqualendrow}[3]{%
  \par\noindent
  \begin{minipage}{\linewidth}
    \centering
    \def\@captype{figure}%
    \setlength{\abovecaptionskip}{4pt}%
    \setlength{\belowcaptionskip}{0pt}%
    \includegraphics[width=\linewidth]{#1}
    \caption{#2}\label{#3}%
  \end{minipage}\par\addvspace{6pt}%
}
\makeatother

\Needspace{200pt}
\section{Supplementary Qualitative Results}
\label{app:qualitative-results}

This appendix presents $24$ additional qualitative examples in four categories: Human, Open, Text, and Scientific. Each pair compares the reference output (Base, left) with a selected RGDR result (right); the description below identifies the correction target for the corresponding pair.

\begingroup
\setlength{\parskip}{0pt}
\appqualrow{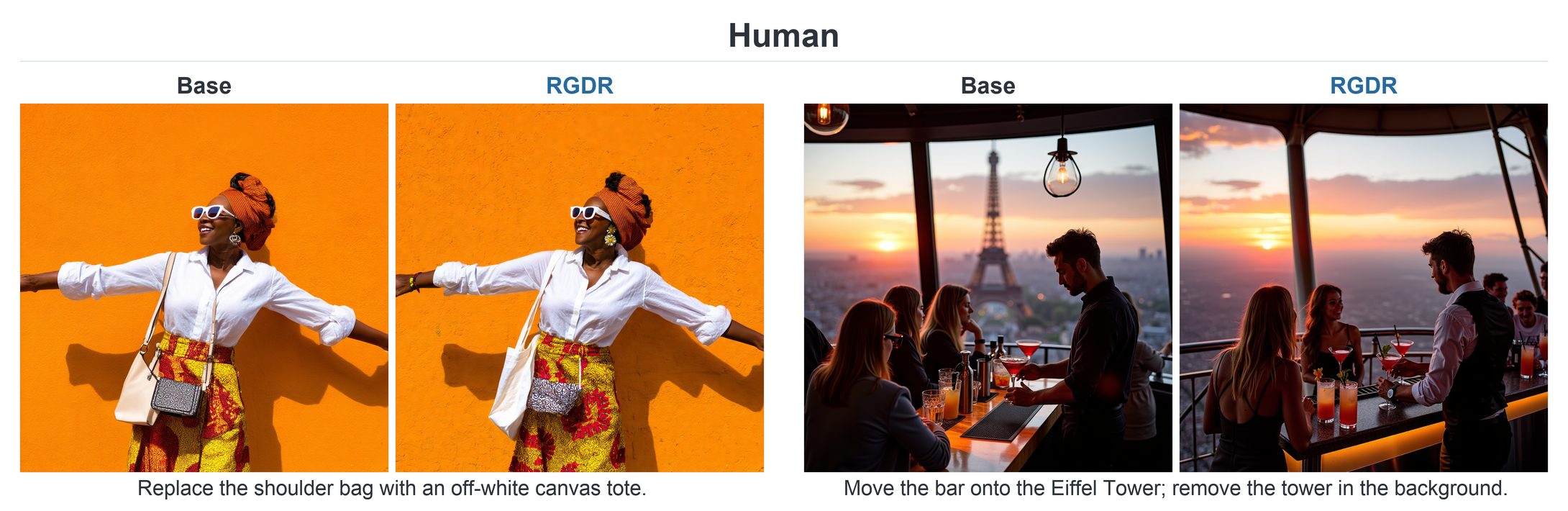}
\appqualrow{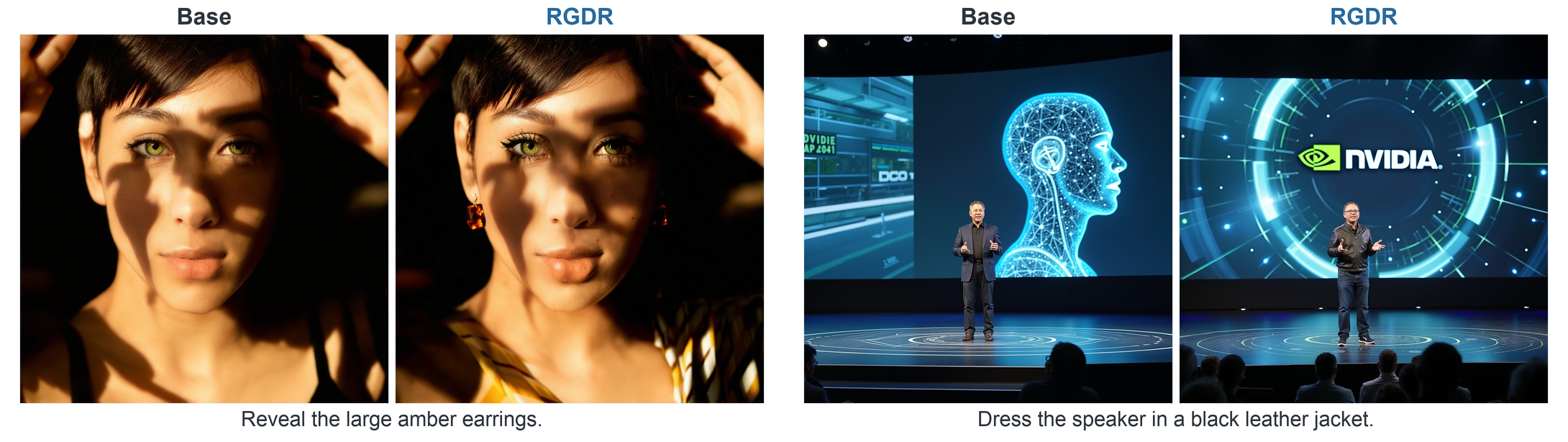}
\appqualendrow{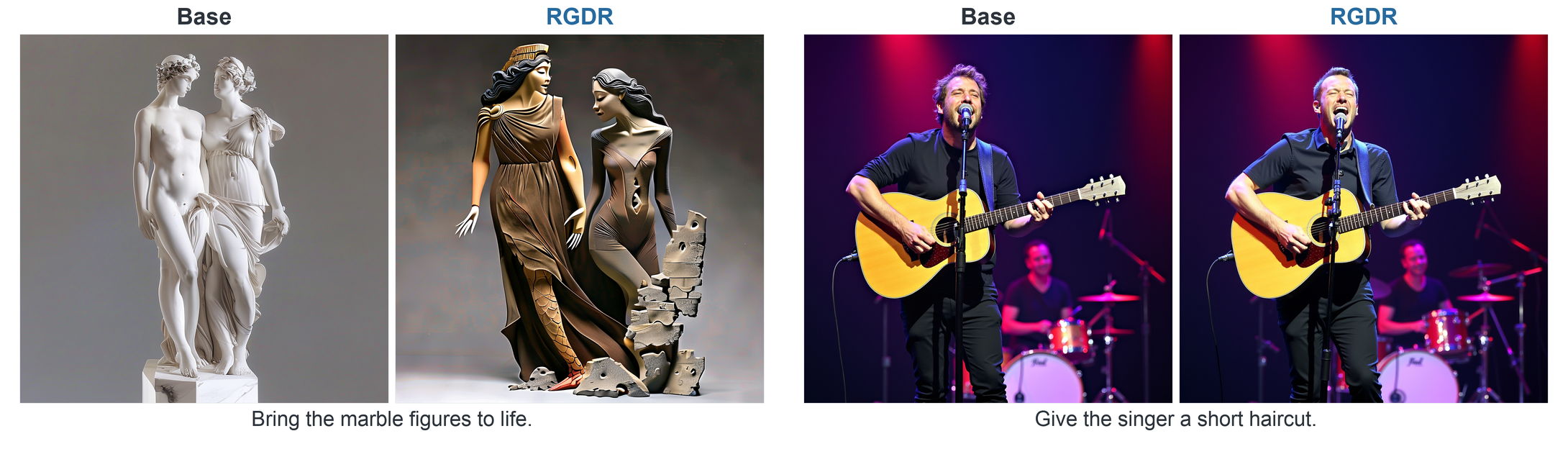}{\textbf{Human.} Base (left) and RGDR (right).}{fig:app-qualitative-human}

\appqualrow{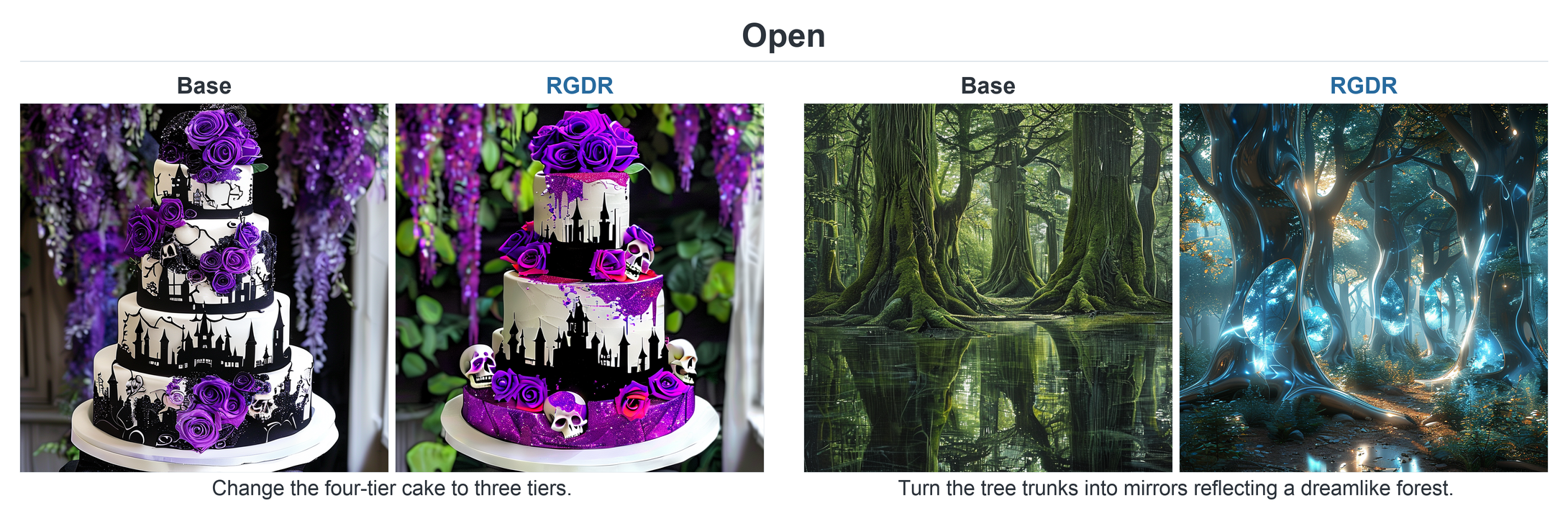}
\appqualrow{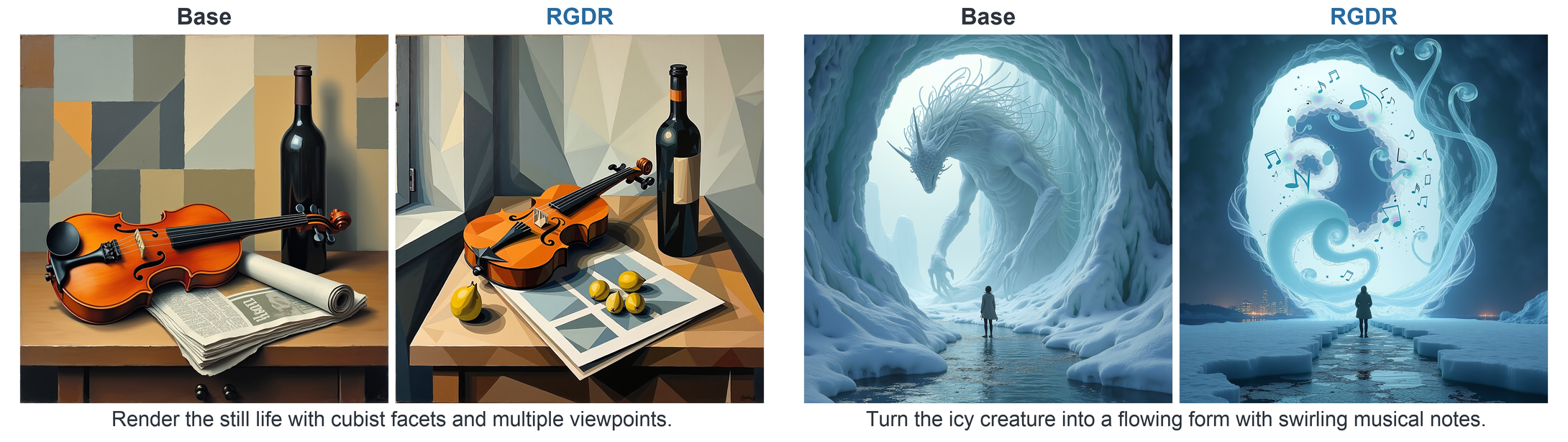}
\appqualendrow{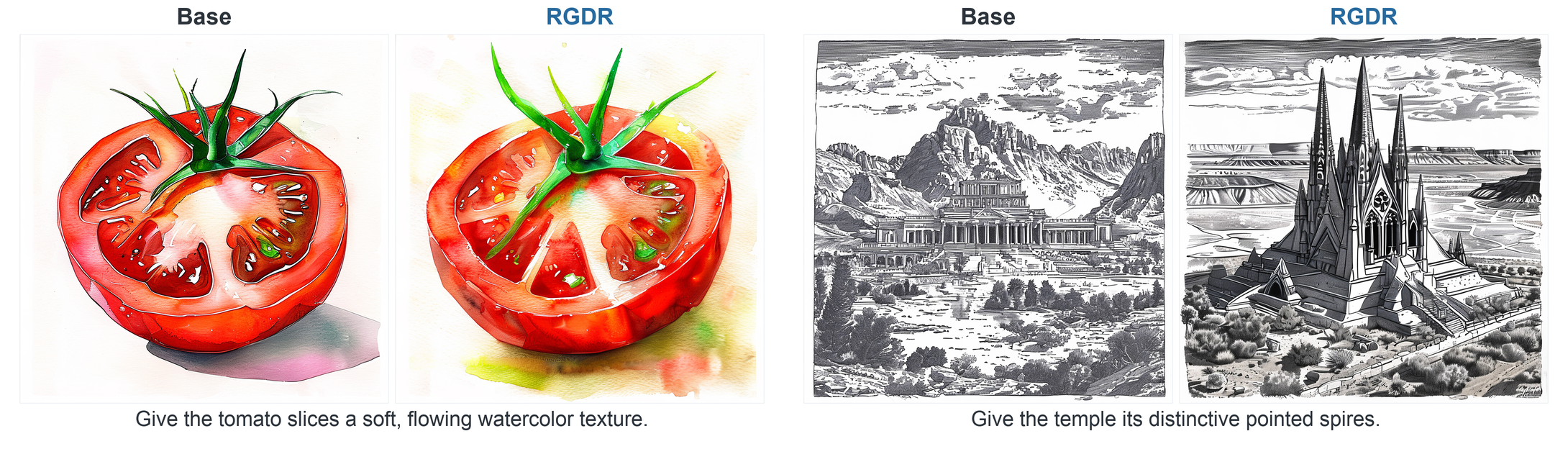}{\textbf{Open.} Base (left) and RGDR (right).}{fig:app-qualitative-open}

\appqualrow{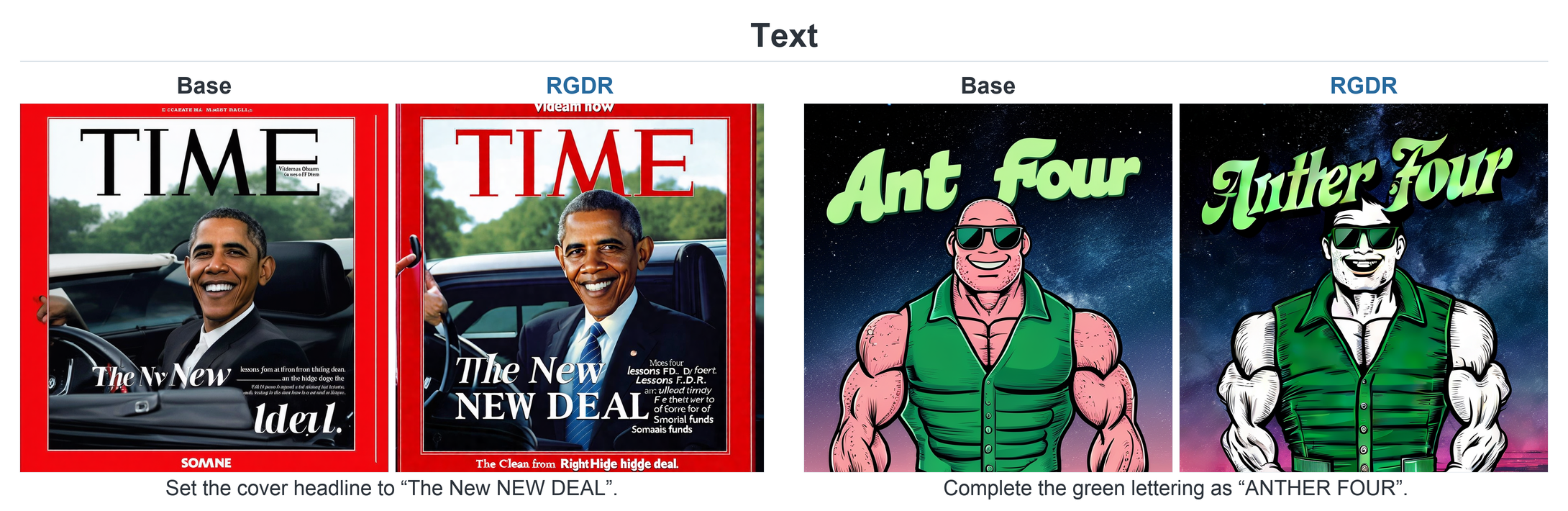}
\appqualrow{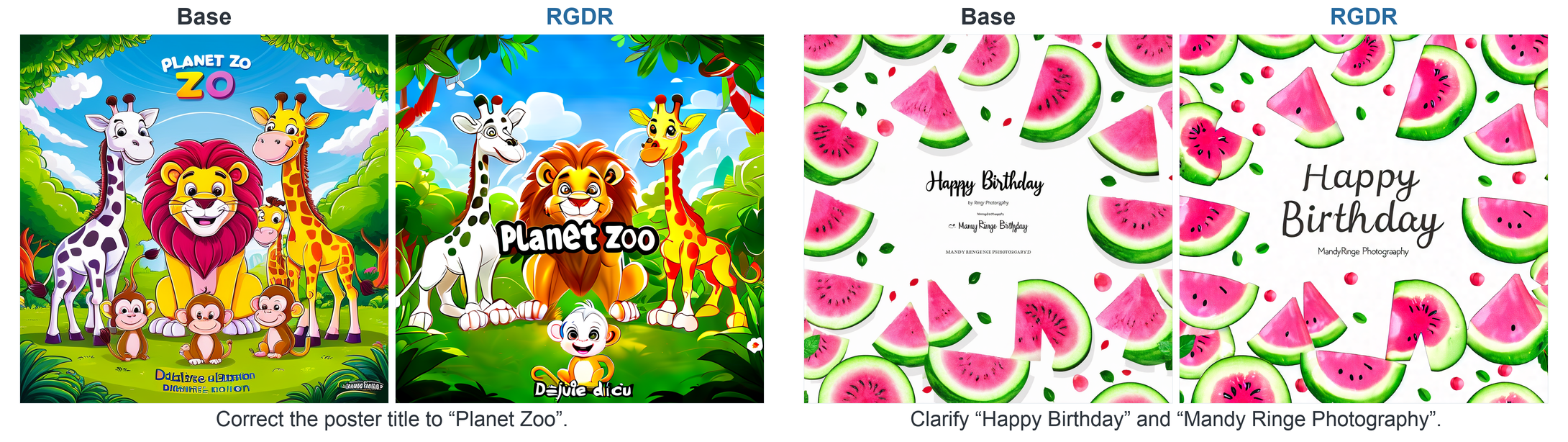}
\appqualendrow{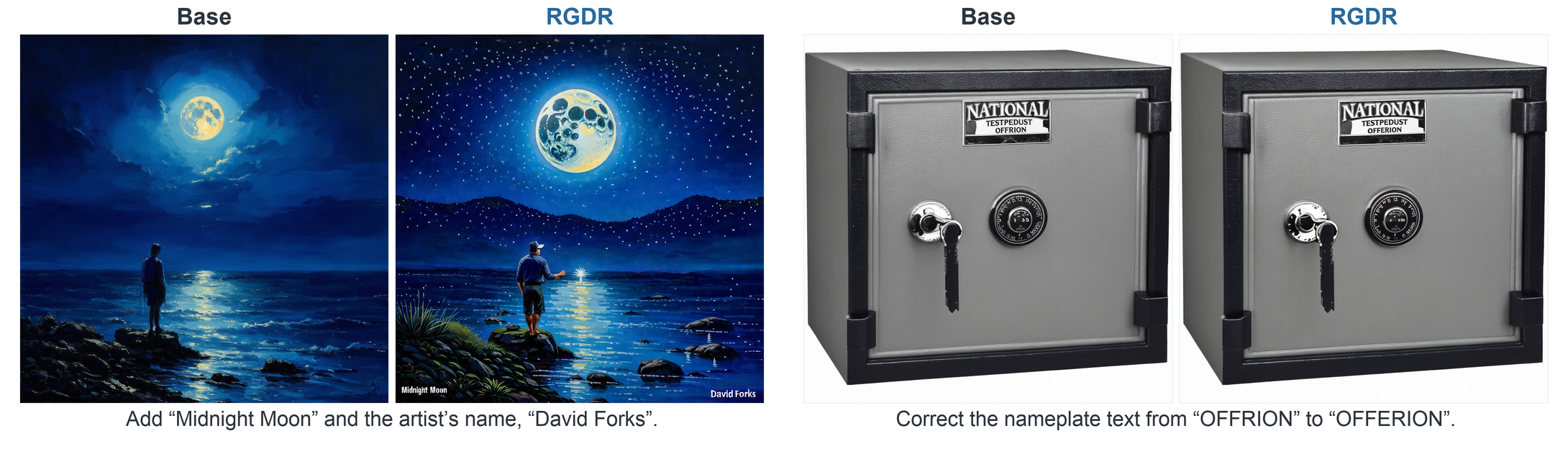}{\textbf{Text.} Base (left) and RGDR (right).}{fig:app-qualitative-text}

\appqualrow{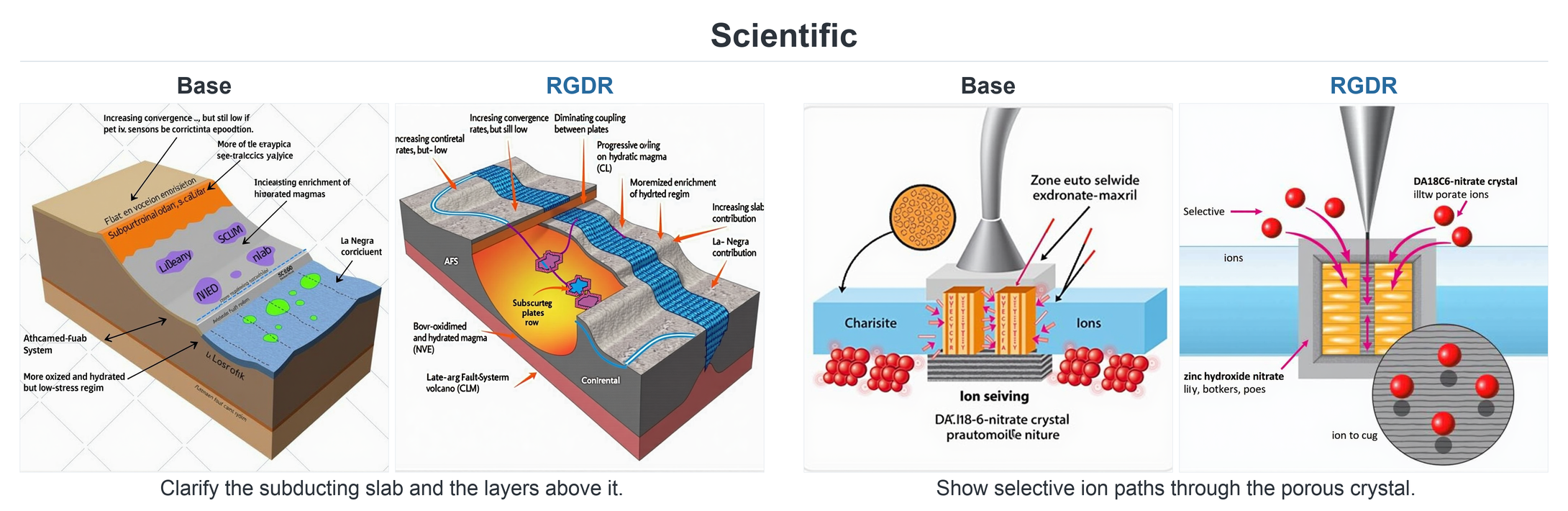}
\appqualrow{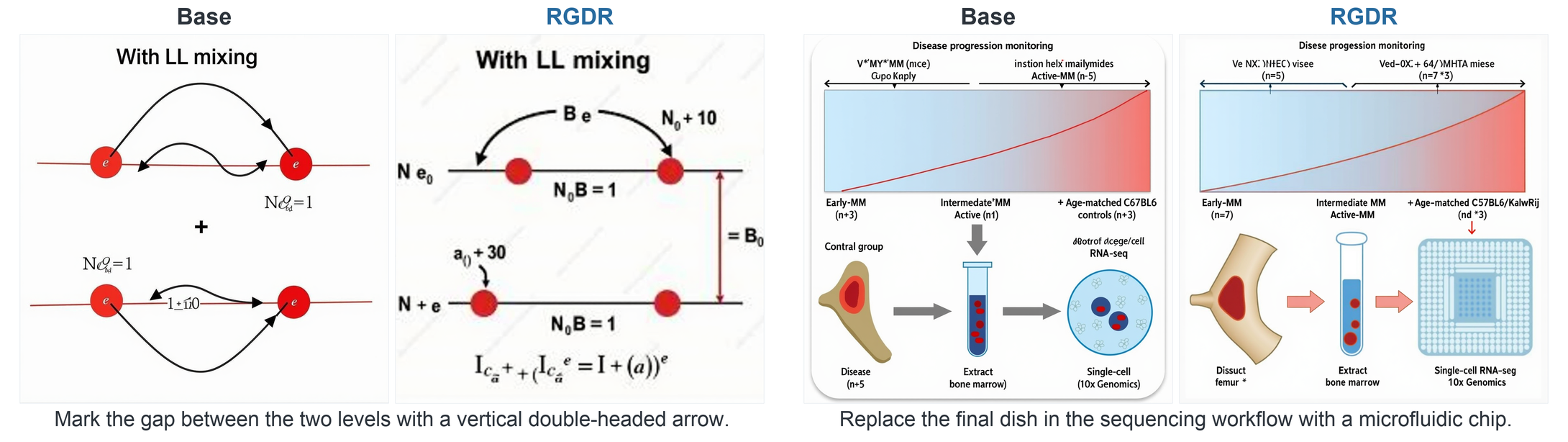}
\appqualendrow{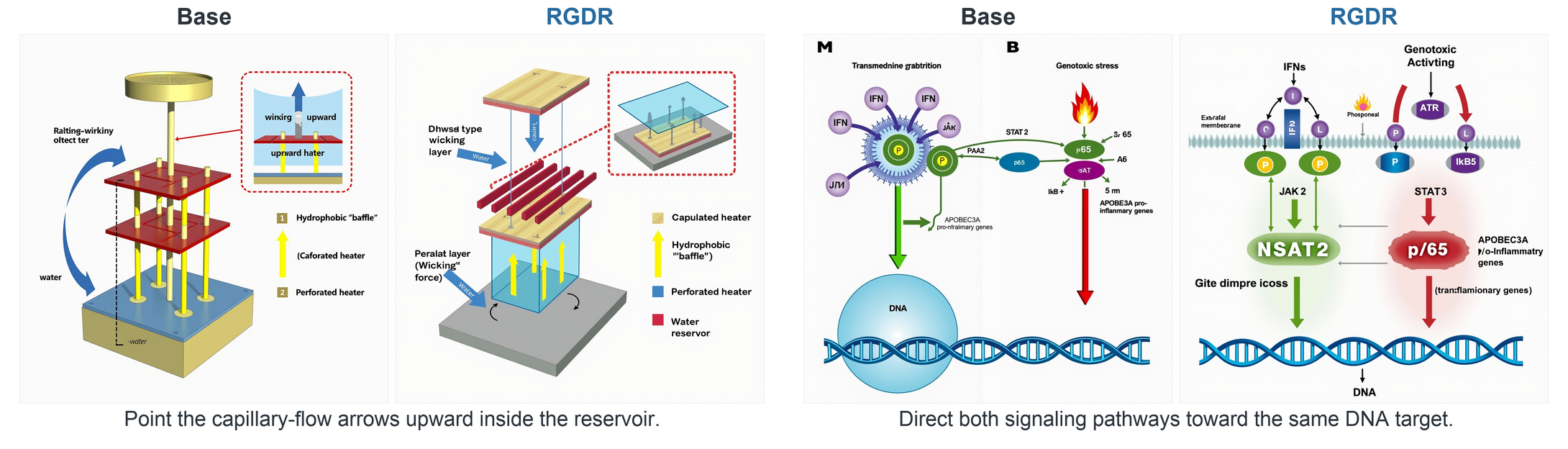}{\textbf{Scientific.} Base (left) and RGDR (right).}{fig:app-qualitative-scientific}
\endgroup

\end{document}